\documentclass[10pt,twocolumn,letterpaper]{article}

\usepackage{cvpr}              

\usepackage{multirow}
\usepackage{enumitem}
\usepackage{booktabs}
\newcommand\textblue{\textcolor[RGB]{3, 111, 193}}
\newcommand\textorange{\textcolor[RGB]{255, 153, 0}}

\definecolor{cvprblue}{rgb}{0.21,0.49,0.74}
\usepackage[pagebackref,breaklinks,colorlinks,allcolors=cvprblue]{hyperref}

\def\paperID{****} 
\def\confName{CVPR}
\def\confYear{2026}

\title{Making Training-Free Diffusion Segmentors Scale with the Generative Power}

\author{\parbox{14cm}{
    \centering
    \large{
        Benyuan Meng$^{1,2}$ \quad
        Qianqian Xu$^{3,8}$\thanks{Corresponding authors.} \quad  
        Zitai Wang$^{3}$
    }
    \\ 
    \large{
        Xiaochun Cao$^{4}$ \quad
        Longtao Huang$^{5}$ \quad
        Qingming Huang$^{6,3,7*}$  
    }
    \\
    \normalsize{
        \textnormal{$^1$Institute of Information Engineering, CAS}\\
        \textnormal{$^2$School of Cyber Security, University of Chinese Academy of Sciences}\\
        \textnormal{$^3$State Key Laboratory of AI Safety, Institute of Computing Technology, CAS}\\
        \textnormal{$^4$School of Cyber Science and Tech., Shenzhen Campus of Sun Yat-sen University}\\
        \textnormal{$^5$Alibaba Group}\\
        \textnormal{$^6$School of Computer Science and Tech., University of Chinese Academy of Sciences}\\
        \textnormal{$^7$Key Laboratory of Big Data Mining and Knowledge Management, CAS}\\
        \textnormal{$^8$Beijing Academy of Artificial Intelligence}
    }
    \\
    \texttt{mengbenyuan@iie.ac.cn}\quad
    \texttt{\{xuqianqian,wangzitai\}@ict.ac.cn}\\  
    \texttt{caoxiaochun@mail.sysu.edu.cn}\quad  
    \texttt{kaiyang.hlt@alibaba-inc.com}\quad
    \texttt{qmhuang@ucas.ac.cn}
}}

\begin{document}
\maketitle

\begin{abstract}
As powerful generative models, text-to-image diffusion models have recently been explored for discriminative tasks.
A line of research focuses on adapting a pre-trained diffusion model to semantic segmentation without any further training, leading to training-free diffusion segmentors.
These methods typically rely on cross-attention maps from the model’s attention layers, which are assumed to capture semantic relationships between image pixels and text tokens.
Ideally, such approaches should benefit from more powerful diffusion models, \textit{i.e.}, stronger generative capability should lead to better segmentation.
However, we observe that existing methods often fail to scale accordingly.
To understand this issue, we identify two underlying gaps: (i) Cross-attention is computed across multiple heads and layers, but there exists a discrepancy between these individual attention maps and a unified global representation. (ii) Even when a global map is available, it does not directly translate to accurate semantic correlation for segmentation, due to score imbalances among different text tokens. To bridge these gaps, we propose two techniques: auto aggregation and per-pixel rescaling, which together enable training-free segmentation to better leverage generative capability.
We evaluate our approach on standard semantic segmentation benchmarks and further integrate it into a generative technique, demonstrating both improved performance and broad applicability.
Codes are at \href{https://github.com/Darkbblue/goca}{https://github.com/Darkbblue/goca}.
\end{abstract}    
\section{Introduction}

Text-to-image diffusion models have become leading generative methods with strong image synthesis capabilities~\cite{ho2020denoising, dhariwal2021guideddiffusion, rombach2022high, podell2023sdxl, han2025lightfair}.
It suggests that diffusion models may effectively understand visual content.
This has inspired research on using them for discrimination~\cite{baranchuk2021label, zhao2023unleashing, xu2023open, zhang2023tale, luo2023dhf, yang2023diffusion, DBLP:conf/cvpr/DuttMM24, frick2024diffseg, my-SDXL, my-GATE, DBLP:conf/cvpr/ZhangYM24}, which can be named \textit{diffusion discriminator}.
It poses strong application potential, as it can benefit from the rapid advancement of generative diffusion models.

A subset of diffusion discriminator research focuses on training-free scenarios, where a diffusion model pre-trained for generation is provided, and no further training is allowed~\cite{xiao2023text, DBLP:journals/corr/abs-2309-02773, DBLP:journals/access/KawanoA24a, DBLP:journals/corr/abs-2502-04320, DBLP:journals/corr/abs-2409-03209, DBLP:conf/aaai/ParkJB25}.
Though quite challenging, such studies can be of importance in various applications, including standalone segmentation tasks~\cite{xiao2023text, DBLP:journals/corr/abs-2309-02773, DBLP:journals/access/KawanoA24a}, model behavior interpretation~\cite{daam}, and integration in advanced generative techniques~\cite{S-CFG, DBLP:conf/iclr/XiaoL0WH24, DBLP:conf/cvpr/Liu0CJH24}.
We term such studies \textit{training-free diffusion segmentors}.
Training-free diffusion segmentors primarily rely on the use of cross-attention maps~\cite{vaswani2017attention}, a special type of sparse feature extracted from intermediate activations during the operation of diffusion models~\cite{baranchuk2021label, xu2023open, my-SDXL}.
Cross-attention maps are believed to reflect the semantic correlation between latent pixels and prompt tokens.
For instance, the prompt token with the highest attention score with respect to one latent pixel is likely to be exactly the semantic category of this pixel.
This property enables segmentation without model retraining.

\begin{figure}[t]
  \centering
  \includegraphics[width=.93\linewidth]{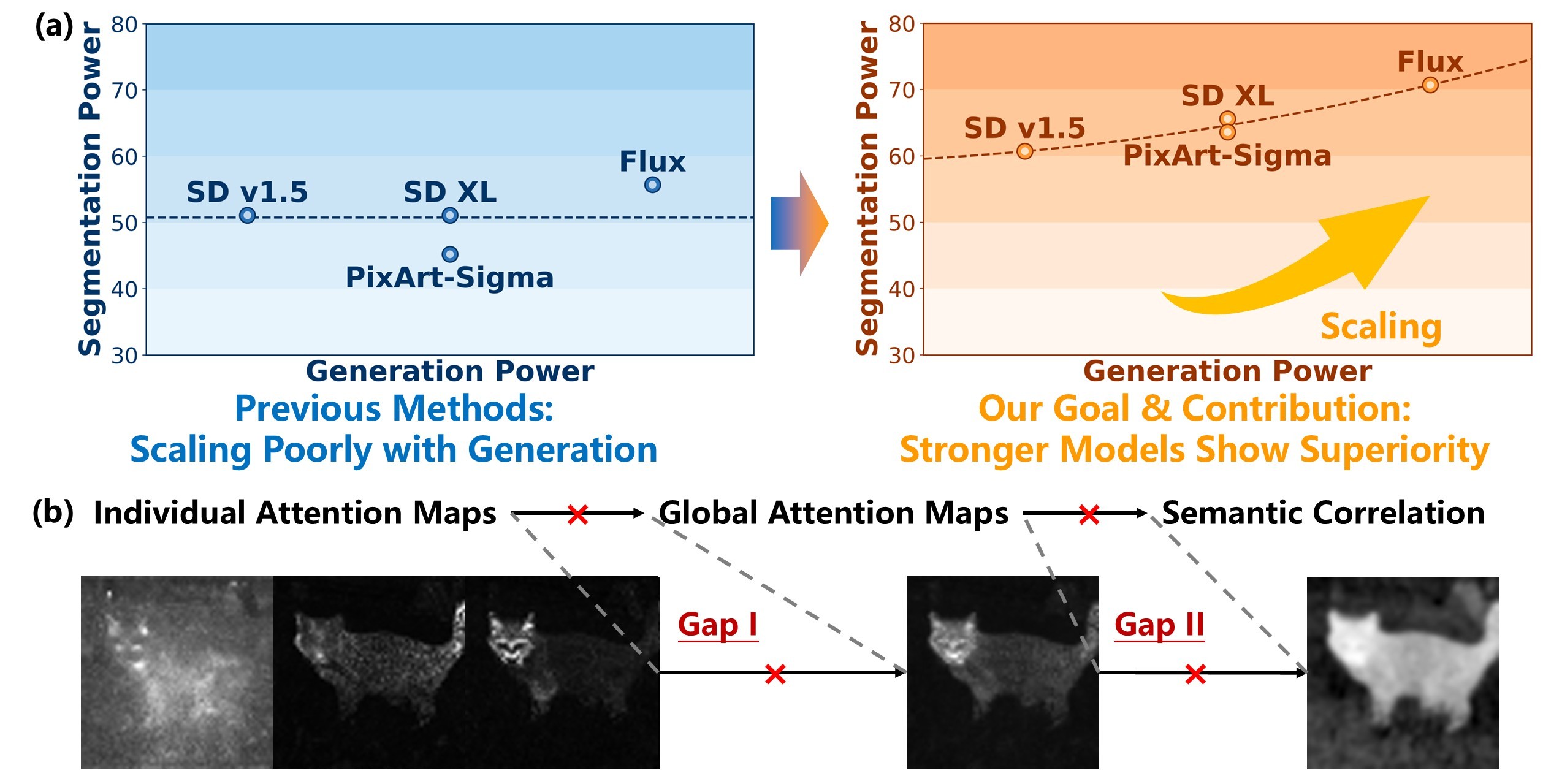}
  \caption{(a) Previous training-free diffusion segmentors scale poorly with the generative power of diffusion models, which inspires our study to enable such scaling.
  (b) We have identified two gaps from individual cross-attention maps to semantic correlation, which have been preventing the aforementioned scaling.
  }
  \label{fig:intro}
\end{figure}

Since training-free diffusion segmentors are based on the generative capability of diffusion models, we expect a stronger model to yield better segmentation results, where the generative power can be evaluated using FID scores~\cite{dowson1982frechet}, CLIP scores~\cite{DBLP:conf/emnlp/HesselHFBC21}, and user study.
However, we notice that most of the previous studies are conducted on very limited diffusion models, \ie, Stable Diffusion v1.5 and Stable Diffusion v2.1~\cite{rombach2022high}, the generative capability of which is relatively outdated.
This limitation raises concerns about their ability to scale to more recent models with much stronger generative capability.
Hence, we conduct an empirical study, experimenting with three stronger diffusion models: Stable Diffusion XL~\cite{podell2023sdxl} (U-Net-based), PixArt-Sigma~\cite{pixart-sigma} (DiT-based), and Flux~\cite{flux2024} (MMDiT-based).
As shown in \Cref{fig:intro}(a), current methods perform poorly on these models, sometimes even yielding worse segmentation, much in the contrary to our expectation.

Why cannot current training-free methods scale well with the generative power of diffusion models?
Specifically, there are many cross-attention heads and layers in diffusion model backbones, each yielding an individual attention map, which needs to be aggregated into a global attention map.
Afterward, this global attention map needs to be interpreted as semantic correlation scores to perform the segmentation.
However, there are two gaps in this two-step process, as shown in \Cref{fig:intro}(b):
(i) Individual attention maps are conventionally aggregated by manually assigning weights~\cite{xiao2023text, DBLP:journals/corr/abs-2309-02773, DBLP:journals/access/KawanoA24a}, which, however, becomes more impractical for more complex architectures in stronger models.
(ii) Global attention scores are not readily the assumed semantic correlation due to score imbalance across tokens.
This gap, in fact, has not been thoroughly addressed yet, but it has not been fully exposed due to the lack of architectural diversity of models validated on.

\textbf{Hence, our goal is to bridge the two gaps and thus enable training-free diffusion segmentors to perform better with stronger diffusion models.}
For the first gap, we have devised \textbf{auto aggregation}, which aggregates individual attention maps according to the correlation among diffusion model activations instead of manual weights, where an individual map with a higher contribution to the overall generation has a higher weight for the segmentation.
For the second gap, we introduce \textbf{per-pixel rescaling} to mitigate interference from semantic special tokens in prompts, drawing global attention maps closer to semantic correlation.
Combined as \textit{Generative scaling of Cross-Attention} (GoCA), our method enables stronger models to easily surpass previous ones.
We validate the method on multiple standard semantic segmentation benchmarks and as the integration of an advanced generation technique, S-CFG~\cite{S-CFG}, showing the effectiveness of GoCA in various scenarios.

In general, our contribution is as follows:
\begin{itemize}[leftmargin=10pt]
    \item To the best of our knowledge, we are the first to point out the inefficacy of current training-free diffusion segmentors to scale with the generative power.
    \item We attribute the inefficacy to the two gaps between cross-attention maps and semantic correlation, and devise auto aggregation and per-pixel rescaling techniques to address the two identified gaps accordingly.
    \item We have validated our method with extensive experiments, including both standard semantic segmentation and the integration of generative techniques.
\end{itemize}

\section{Related Work}
To enable text to steer image generation, Stable Diffusion introduces cross-attention modules between image and text~\cite{rombach2022high}.
Then, cross-attention maps extracted from such modules were found to be able to explain certain behaviors of text-to-image diffusion models~\cite{daam}:
a text token with higher attention scores with respect to a pixel contributes more to the generation of this pixel.
Later, this discovery was extended from a model diagnosis tool to direct application in semantic segmentation, namely training-free diffusion segmentors, as they require no additional training~\cite{xiao2023text, DBLP:journals/corr/abs-2309-02773, DBLP:journals/access/KawanoA24a, DBLP:journals/corr/abs-2502-04320, DBLP:journals/corr/abs-2409-03209, DBLP:conf/aaai/ParkJB25}.
It soon became a common practice in this field to manually aggregate attention maps from different layers, and the focus of studies turned to the post-processing or further refinement of attention maps.
For example, using self-attention maps for refinement (spatial propagation) is shown to be very effective~\cite{xiao2023text, DBLP:journals/access/KawanoA24a}.
Additionally, there are also prototype-based refinement~\cite{DBLP:journals/access/KawanoA24a} and dCRF-based refinement~\cite{xiao2023text} proposed.

However, previous studies have been heavily relying on Stable Diffusion v1.5 as the base model, and this poses a potential threat to the generalizability of the proposed methods over model architecture.
Our work taps into this topic.

\section{Preliminary}
\label{sec:pre}
\subsection{Diffusion Model}
Diffusion models generate images by progressively removing noises from a partially noisy image~\cite{ho2020denoising}.
Training samples can be acquired by adding noises $\epsilon$ to clean images $x_0$:
\begin{equation}
    \label{equ:diffusion-1}
    q(x_t|x_{t-1}):=\mathcal{N}(x_t;\sqrt{1-\beta_t}x_{t-1},\beta_tI).
\end{equation}
where $\beta_1, \cdots, \beta_t$ are a group of variance constants, and $t$ indicates noise strength.
A diffusion model contains a neural network $\epsilon_\theta$, as parameterized by $\theta$, trained to predict the noises to remove at each step.
Once the network is trained on vast image samples, we can predict noises as
\begin{equation}
    \label{equ:diffusion-3}
    x_{t-1}=\frac{1}{\sqrt{\alpha_t}}(x_t-\frac{1-\alpha_t}{\sqrt{1-\overline{\alpha}}_t}\epsilon_\theta(x_t,t))+\sigma_t\epsilon.
\end{equation}
where $\epsilon\sim\mathcal{N}(0,I)$, $\alpha_t:=1-\beta_t$, $\overline{\alpha}_t:=\prod_{s=1}^t\alpha_s$, and $\sigma^2=\beta_t$.

\subsection{Attention Mechanism}
Training-free diffusion segmentors rely on the cross-attention maps extracted from cross-attention layers in a diffusion model's noise predictor network.
Such cross-attention layers are a variant of typical attention layers~\cite{vaswani2017attention}.

In a typical attention layer, there are three input sequences: query ($\mathbf{Q}\in\mathbb{R}^{n_q\times d_q}$), key ($\mathbf{K}\in\mathbb{R}^{n_k\times d_k}$), and value ($\mathbf{V}\in\mathbb{R}^{n_v\times d_v}$), where $n_q, n_k, n_v$ are sequence length, and $d_q, d_k, d_v$ are embedding channels.
Usually, we have $n_k=n_v$ and $d_q=d_k$.
The three sequences are first linearly projected with $\mathbf{W}^Q\in\mathbb{R}^{d_q\times d_q}, \mathbf{W}^K\in\mathbb{R}^{d_k\times d_k}, \mathbf{W}^V\in\mathbb{R}^{d_v\times d_v}$, resulting in $\mathbf{QW}^Q\in\mathbb{R}^{n_q\times d_q}, \mathbf{KW}^K\in\mathbb{R}^{n_k\times d_k}, \mathbf{VW}^V\in\mathbb{R}^{n_v\times d_v}$.
Afterward, we compute the dot-product similarity between query and key as
\begin{equation}
    \label{equ:attn}
    \mathbf{A} = Softmax(\frac{(\mathbf{QW}^Q)(\mathbf{KW}^K)^T}{\sqrt{d_q}}) \in\mathbb{R}^{n_q\times n_k}.
\end{equation}
\textbf{Notably, this} $\mathbf{A}$ \textbf{matrix in \Cref{equ:attn} is the attention map commonly referenced in training-free diffusion segmentors.}
Afterward, $\mathbf{A}$ serves as a weight matrix, leading to a weighted average over tokens in the value sequence ($\mathbf{A}(\mathbf{VW}^V)$).
Finally, the averaged result is passed through an extra projection $\mathbf{W}^O\in\mathbb{R}^{d_v\times d_q}$, \ie, $\mathbf{Q}\leftarrow \mathbf{Q}+\mathbf{A}(\mathbf{VW}^V)\mathbf{W}^O$.

Typically, the attention layers in diffusion models use multi-head attention.
Taking head count $h$ as an example, each sequence is projected into $h$ sequences.
Then, $\mathbf{A}$ is computed for each separate sequence, which leads to weighted average results for each of them.
Finally, these results are concatenated together before being projected through $\mathbf{W}^O$:
\begin{equation}
    \label{equ:multi-head}
    Output=Concat(\mathbf{A}_1\mathbf{V}_1,\cdots,\mathbf{A}_h\mathbf{V}_h)\mathbf{W}^O.
\end{equation}

Specifically, in cross-attention layers of diffusion models, query sequence is the latent pixels, and key/value sequences are prompt embeddings.
Thus, it is believed that the cross-attention maps can be regarded as the semantic correlation between latent pixels and prompt tokens.

\section{Problem Analysis \& Method}

\begin{figure}[t]
  \centering
  \includegraphics[width=.9\linewidth]{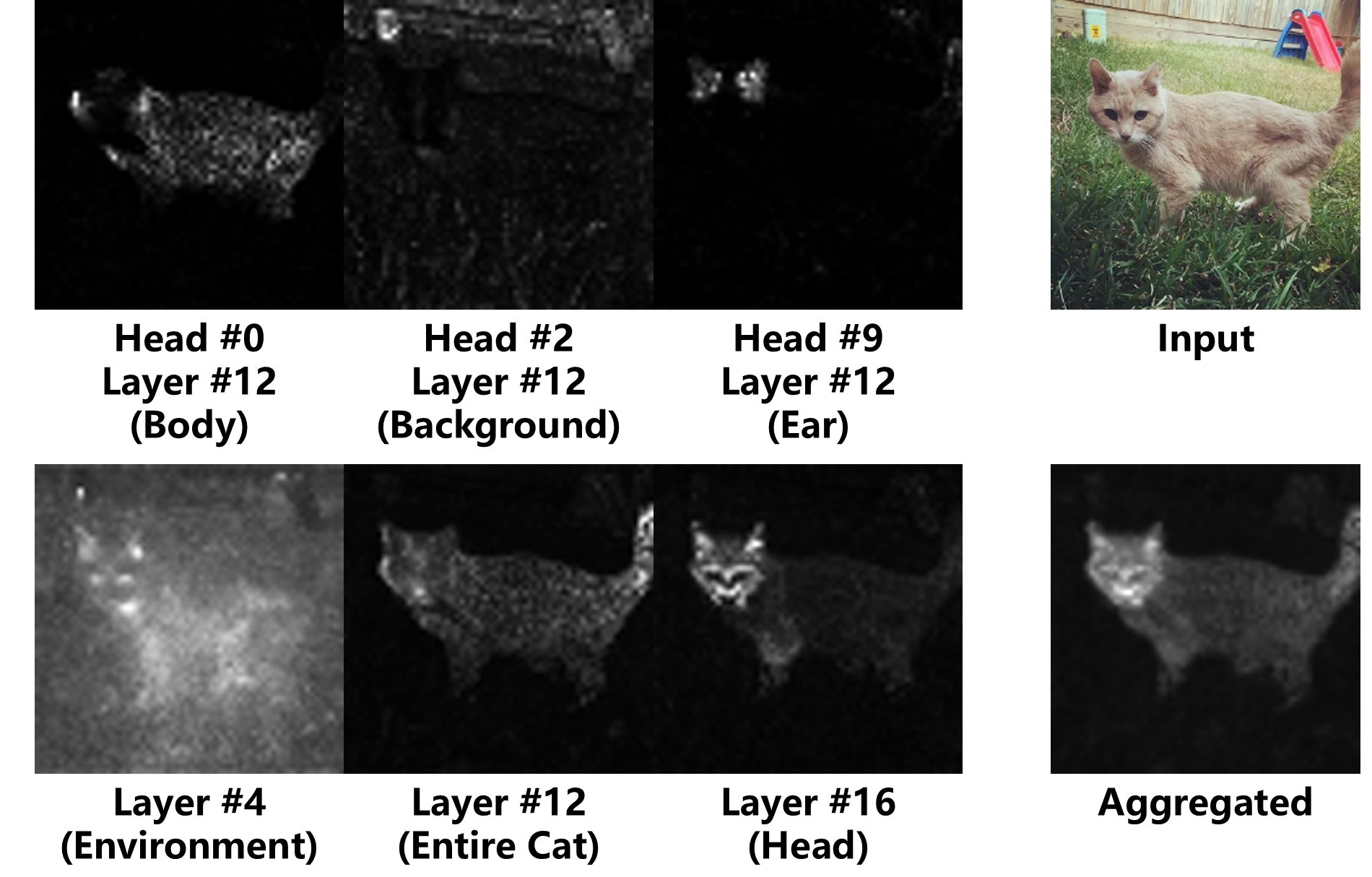}
  \caption{Attention maps in different heads and layers show a certain collaboration pattern, each focusing on distinct aspects of the image.
  }
  \label{fig:method-1}
\end{figure}

\begin{figure*}[t]
  \centering
  \includegraphics[width=.85\linewidth]{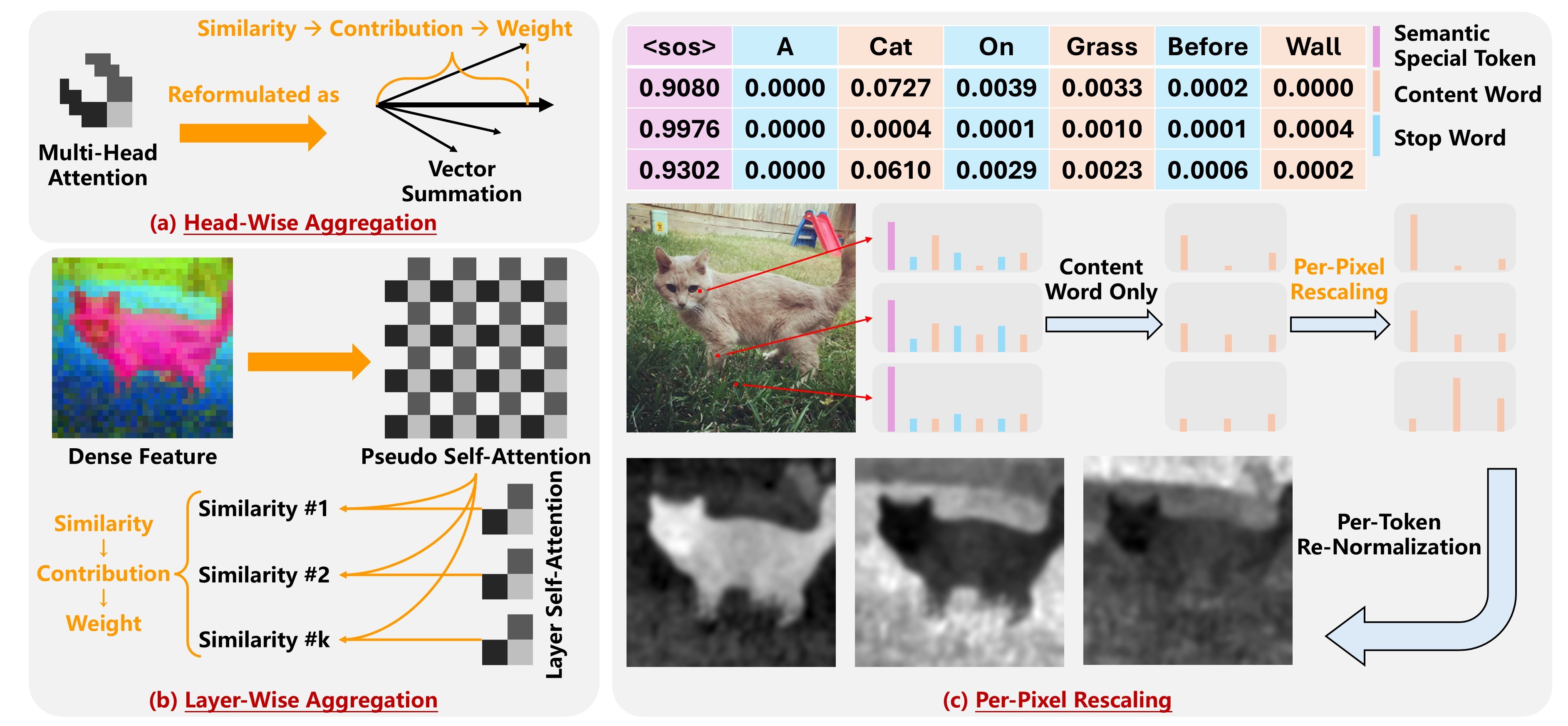}
  \caption{Overview of our method. (a) and (b) consist the auto aggregation part, and (c) is the per-pixel rescaling part.
  (a) Head-wise aggregation reformulates multi-head attention as vector summation and then uses the dot-product similarity between each vector and the summed vector as head weights.
  (b) Layer-wise aggregation computes pseudo self-attention based on a chosen dense feature, regards it as the pseudo global attention, and finally uses the similarity between per-layer self-attention maps and this pseudo global attention as layer weights.
  (c) We exclude semantic special tokens and stop word tokens, only considering content word tokens, to rescale their attention scores to sum to 1, followed by a conventional per-token re-normalization.
  }
  \label{fig:method}
\end{figure*}
The entire method overview is depicted in \Cref{fig:method}.
In the following, we will elaborate on the problem we are facing before offering our solution both intuitively and formally.

\subsection{Addressing Gap I: Auto Aggregation}

\textbf{Problem Statement}.
As shown in \Cref{fig:method-1}, collaboration patterns emerge among attention heads and layers.
Their combined effect, which leads to the generated image, is believed to reflect overall affinity.
However, the individual maps focus on each head's and layer's own jobs, creating a gap between individual attention scores and a unified map.
Previous training-free diffusion segmentors aggregate multiple cross-attention maps using a weighted average, where the weights are manually tuned~\cite{xiao2023text, DBLP:journals/corr/abs-2309-02773, DBLP:journals/access/KawanoA24a}.
This manual practice struggles with stronger diffusion models, as they tend to have more complex architectures, making manual tuning difficult.
\textbf{Therefore, our goal is to design a technique to automatically set the aggregation weights based on each map's contribution to the overall generation, derived from relationships of model activations.}

\textbf{Head-Wise Aggregation}.
Intuitively, as shown in \Cref{fig:method}(a), the collaboration of different heads in a multi-head attention mechanism can be reformulated into vector summation~\cite{DBLP:journals/corr/abs-2410-11842}, which will be shown formally later.
In this way, each head's relative contribution to the layer output relates to the similarity between its vector and the summed output vector~\cite{DBLP:journals/corr/abs-2410-11842}.
Then, we can base our auto weights on this relative contribution, where a head with a higher contribution gets a higher weight.

Formally, one multi-head attention layer defined in \Cref{equ:multi-head} can be reformulated as vector summation:
\begin{equation}
    \label{equ:head-wise}
    Output=\mathbf{A}_1\mathbf{V}_1\mathbf{W}_1^O+\cdots+\mathbf{A}_h\mathbf{V}_h\mathbf{W}_h^O.
\end{equation}
where $[(\mathbf{W}_1^O)^T|\cdots|(\mathbf{W}_h^O)^T]^T=\mathbf{W}^O$.
Afterward, with $m$ as layer index and $n$ as head index, we can compute the contribution $w_{mn}$ of each individual vector $(\mathbf{A}_n\mathbf{V}_n\mathbf{W}_n^O)_m$ using dot-product similarity:
\begin{equation}
    w_{mn} = (\mathbf{A}_n\mathbf{V}_n\mathbf{W}_n^O)_m^TOutput_m.
\end{equation}
After normalizing the head contribution ($w_{mn}' = \frac{w_{mn}}{\sum_{j=1}^hw_{mj}}$), we obtain head-wise weights $w_{m1}',\cdots, w_{mh}'$.
In addition, the process above is done in parallel for each pixel, which means we have per-pixel weights instead of one overall weight for the entire attention map.
The aggregation of per-head maps is then done as
\begin{equation}
    \label{equ:head-wise-aggregation}
    \mathbf{A}_{m}=\sum_{n=1}^h w_{mn}'\mathbf{A}_{mn}.
\end{equation}
As such, we can aggregate maps of different heads within one layer, obtaining per-layer attention maps.
With this ready, we will be doing the layer-wise aggregation.

\textbf{Layer-Wise Aggregation}.
Similarly, we wish to compute the contribution of each layer to the overall generation.
Here, we measure the contribution with the similarity between each individual attention map and a global map.
As we do not have access to the global attention map yet, we instead need a proxy.
Specifically, as shown in \Cref{fig:method}(b), we can compute spatial affinity from dense diffusion features~\cite{baranchuk2021label, my-SDXL}, which can be used as pseudo self-attention maps~\cite{DBLP:journals/corr/abs-2406-02842, DBLP:journals/access/KawanoA24a}.
More importantly, they are close to the global attention scores~\cite{DBLP:journals/corr/abs-2406-02842, DBLP:journals/access/KawanoA24a}.
Therefore, such pseudo self-attention maps can serve as the proxy.
Notably, this pseudo global map is self-attention instead of cross-attention.
Hence, we further hypothesize that the contribution of cross-attention layers resembles that of self-attention layers.
In this way, the aggregation weights for cross-attention layers are obtained by computing the similarity between per-layer self-attention maps and the pseudo self-attention map.

Formally, we denote the dense feature as $feat\in\mathbb{R}^{hw\times d}$, where $d$ is the channel of the dense feature, and $hw$ is the product of the latent image's height and width.
This dense feature can be obtained by selecting a high-quality feature from a relatively high-resolution layer, based on the quantitative results reported in a recent dense diffusion feature study~\cite{my-SDXL}.
Then, the pseudo self-attention is computed as
\begin{equation}
    \label{equ:pseudo-self-attn}
    \mathbf{A}_{p} = Softmax(\frac{feat(feat)^T}{\sqrt{d}}) \in\mathbb{R}^{hw\times hw}.
\end{equation}
Other regular self-attention maps are denoted as $\mathbf{A}_{self}\in\mathbb{R}^{hw\times hw}$.
We reshape these attention maps into $A_{p}', A_{self}' \in\mathbb{R}^{h^2w^2 \times 1}$.
Finally, the weight $w_m$ of layer $m$ is computed as its dot-product similarity with the pseudo self-attention.
\begin{equation}
\begin{aligned}
    \label{equ:layer-wise}
    w_m & = (A_{p}')^T(A^m_{self})'\in\mathbb{R},\\
    w_m'& = \frac{w_m}{\sum_{j=1}^kw_j}.
\end{aligned}
\end{equation}
where $k$ is the number of all layers.
Finally, the aggregation of per-layer maps is done as
\begin{equation}
    \label{equ:layer-wise-aggregation}
    \mathbf{A}=\sum_{m=1}^k w_m'\mathbf{A}_{m}.
\end{equation}
where $\mathbf{A}_m$ is the output of \Cref{equ:head-wise-aggregation}.
Now we have obtained a global attention map $\mathbf{A}$, but it still requires further refinement to become the semantic correlation.

\subsection{Addressing Gap II: Per-Pixel Rescaling}
\begin{figure}[t]
  \centering
  \includegraphics[width=.9\linewidth]{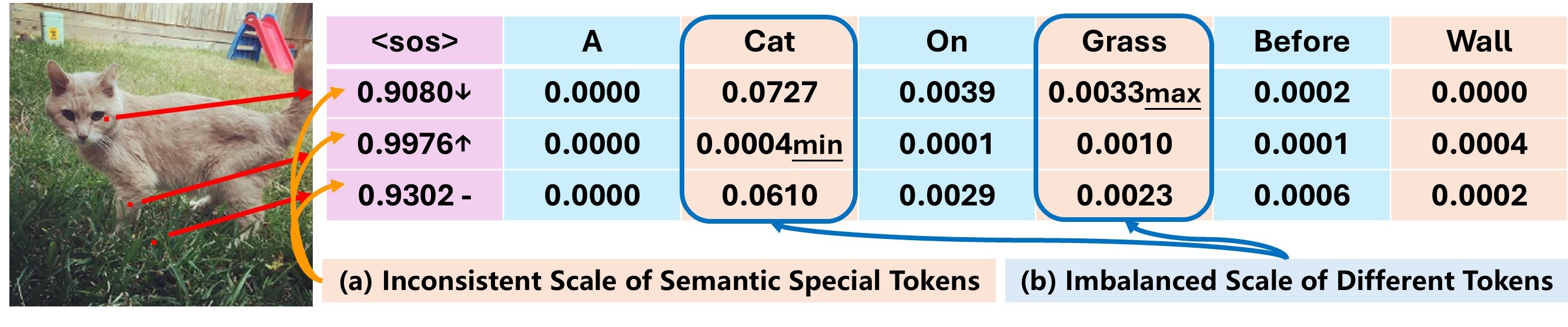}
  \caption{Illustration of imbalance phenomena in raw global attention scores.
  }
  \label{fig:method-2}
\end{figure}

\textbf{Problem Statement}.
There are three types of tokens that comprise a token sequence:
\begin{enumerate}[label=(\roman*), leftmargin=25pt]
    \item Semantic special token, such as ``\textless sos\textgreater'' in Stable Diffusion v1.5~\cite{rombach2022high}, that encodes the entire sentence.
    \item Content word tokens, such as ``cat'' and ``grass'', which have actual meanings and are the focus for training-free segmentation.
    \item Stop word tokens, such as ``of'', ``a'', comma, and padding, which do not have actual meanings.
\end{enumerate}
Training-free diffusion segmentors rely on content word tokens' attention scores: the token with the highest attention score with respect to a pixel is taken as the semantic class of this pixel.
However, shown in \Cref{fig:method-2}, two imbalance phenomena in attention maps are noticed, preventing us from directly retrieving the desired semantic correlation:

(i) \textbf{The scores of different content tokens are imbalanced.}
Usually, foreground objects such as ``cat'' tend to have much higher attention scores than background objects such as ``grass''.
For example, in \Cref{fig:method-2}, even the lowest attention score of ``cat'' is almost comparable to the highest score of ``grass''.
As a result, directly comparing the raw scores of different tokens with respect to one pixel is not reliable.
This problem has been noticed by previous training-free diffusion segmentor studies~\cite{xiao2023text, DBLP:journals/corr/abs-2309-02773, DBLP:journals/access/KawanoA24a}, and they address it by per-token re-normalization, basically normalizing the scores across all pixels with respect to one token.
However, note that this solution implicitly introduces the requirement of cross-pixel score comparison, while previously, only cross-token comparison was needed.

(ii) \textbf{The scores of semantic special tokens and other tokens are imbalanced, and the score scale of semantic special tokens is inconsistent}.
The semantic special token almost always gets much higher attention scores than other tokens.
For some diffusion models, this might be because this token is specially trained to capture the overall semantics of the whole sentence, and for the others, a possible explanation is attention sink~\cite{DBLP:conf/iclr/XiaoTCHL24, DBLP:journals/corr/abs-2505-06708}.
Therefore, a slight change in the scale of the semantic special token's score will significantly change the summed scale of other tokens.
Hence, the direct cross-pixel score comparison, which is required by the solution to the first imbalance, is not reliable either.
For example, in \Cref{fig:method-2}, the first pixel is in the cat's face, and the third pixel is on the grass, so we expect the third pixel to have a higher score for ``grass'' token than the first pixel.
However, as the semantic special token score is lower in pixel one, the score for ``grass'' is made higher here, actually surpassing the ``grass'' score in pixel three.
In such a case, the per-token re-normalization will produce an incorrect result.
\textbf{This problem has not been fully exposed before, and will be our focus next.}

\begin{table*}[t]
\centering
\caption{Experimental results on standard semantic segmentation benchmarks, evaluated using mIoU$\uparrow$.
The best and runner-up are \textorange{\textbf{bold}} and \textblue{\underline{underlined}}.
$^1$Reproduced by us because the original setting is different, details in \Cref{sec:app-reproduction}.}
\label{tab:exp-benchmark}
\begin{tabular}{@{}llccccc@{}}
\toprule
Type                       & Method        & VOC           & Context       & COCO-Object   & Cityscapes    & ADE20K        \\ \midrule
\multirow{2}{*}{Non-DM}    & MaskCLIP      & 38.8          & 23.6          & 20.6          & 10.0          & 9.8           \\
                           & ReCO          & 25.1          & 19.9          & 15.7          & 19.3          & 11.2          \\ \midrule
\multirow{3}{*}{Pre-Trained DM} & DiffSegmentor & 60.1          & 27.5          & 37.9          & -             & -             \\
                           & MaskDiffusion & 29.9          & -             & -             & 17.1          & -             \\
                           & FTTM$^1$          & 48.9          & 30.0          & 34.6          & 12.3          & 20.3          \\ \midrule
\multirow{4}{*}{Vanilla}   & SD v1.5       & 44.3          & 32.3          & 32.3          & 11.8          & 18.0          \\
                           & SD XL         & 51.1          & 35.7          & 37.2          & 16.1          & 18.6          \\
                           & Pixart-Sigma  & 45.2          & 37.0          & 33.4          & 22.5          & 19.1          \\
                           & Flux          & 55.7          & \textblue{\underline{48.4}}    & 43.3         & \textblue{\underline{25.6}}    & \textblue{\underline{24.5}}    \\ \midrule
Baseline                   & SD v1.5       & 51.1          & 35.4          & 36.9          & 18.4          & 21.0          \\ \midrule
\multirow{4}{*}{Ours}      & SD v1.5       & 60.7          & 40.4          & 39.2          & 16.1          & 22.0          \\
                           & SD XL         & \textblue{\underline{65.6}}    & 42.3          & \textblue{\underline{44.3}}          & 21.2          & 23.2          \\
                           & Pixart-Sigma  & 63.6          & 43.2          & 39.8          & 22.6          & 23.8          \\
                           & Flux          & \textorange{\textbf{70.7}} & \textorange{\textbf{51.1}} & \textorange{\textbf{48.1}} & \textorange{\textbf{27.1}} & \textorange{\textbf{29.3}} \\ \bottomrule
\end{tabular}
\end{table*}


\textbf{Solution}.
Intuitively, as shown in \Cref{fig:method}(c), to alleviate the imbalance caused by the semantic special token, we can simply exclude it from the token sequence and rescale the attention scores among the remaining tokens.

Formally, given the scores $\mathbf{A}(i,:)\in\mathbb{R}^n$ of each token with respect to pixel $i$, where $n$ is the number of tokens, we now use per-pixel rescaling to refine the scores.
We grab all content word tokens and ignore the other tokens.
Denoting the indices of content word tokens as $\boldsymbol{q}\in\mathbb{Z}^{l}$, where $l$ is the number of content words in the given prompt, we have $\mathbf{A}(i,\boldsymbol{q})\in\mathbb{R}^l$.
Then we compute the relative scores among content word token scores, normalized to 1:
\begin{equation}
    \label{equ:rescaling}
    \mathbf{A}'(i,\boldsymbol{q})=\frac{\mathbf{A}(i,\boldsymbol{q})}{\sum_{j=1}^l\mathbf{A}(i,\boldsymbol{q}(j))}.
\end{equation}
where $j$ points to each content word token in the token sequence.
After this step, we have filtered out most of the influence of the semantic special token score scale, making cross-pixel score comparison more reliable.

After per-pixel rescaling, we also conduct per-token re-normalization.
Now we are given the attention scores of one token $j$ across all latent pixels as $\mathbf{A}'(\cdot,j)\in\mathbb{R}^{hw}$, where $hw$ is the product of the latent image's height and width.
The re-normalization is done as
\begin{equation}
    \label{equ:renorm}
    \mathbf{A}''(i,j)=\frac{\mathbf{A}'(i,j)-\mathop{min}\limits_{i'}(\mathbf{A}'(i',j))}{\mathop{max}\limits_{i'}(\mathbf{A}'(i',j))-\mathop{min}\limits_{i'}(\mathbf{A}'(i',j))}.
\end{equation}
This step re-normalizes the scores of different tokens to $[0, 1]$, making cross-token comparison more reliable.

After both per-pixel rescaling and per-token re-normalization, the refined attention scores are closer to semantic correlation.
Following previous training-free diffusion segmentors~\cite{DBLP:journals/corr/abs-2309-02773, xiao2023text}, we multiply the final refined cross-attention map $\mathbf{A}''$ by self-attention maps as additional postprocessing refinement.
Finally, we can directly obtain segmentation results without training:
\begin{equation}
    \label{equ:get-label}
    y_i=\mathop{argmax}_j\mathbf{A}''(i,j).
\end{equation}
where $y_i$ is the class label of pixel $i$.

\section{Experiment}
\subsection{Settings}
\textbf{Tasks.}
Semantic segmentation~\cite{hao2020brief, han2024aucseg}, which can be seen as pixel-level classification.

\textbf{Datasets.}
Standard semantic segmentation benchmarks: Pascal VOC 2012~\cite{pascal-voc-2012}, Pascal-Context-59~\cite{pascal-context}, COCO-Object~\cite{coco-object}, Cityscapes~\cite{Cordts16city}, and ADE20K~\cite{Zhou17ade20k}.

\textbf{Metrics.}
For standard segmentation benchmarks, we use mIoU.
It is the mean over the IoU performance across all semantic classes \cite{hao2020brief}.
For each image, IoU (Intersection over Union, $\uparrow$) is defined by \#(overlapped pixels between prediction and ground truth) / \#(union pixels of them).

\textbf{Diffusion Models.}
We experiment with four diffusion models: the relatively outdated Stable Diffusion v1.5~\cite{rombach2022high} (SD v1.5) and three stronger models: Stable Diffusion XL~\cite{podell2023sdxl} (SD XL), PixArt-Sigma~\cite{pixart-sigma}, and Flux~\cite{flux2024}.

\textbf{Evaluated Methods.}
\begin{enumerate}[label=(\roman*), leftmargin=25pt]
    \item \textit{Vanilla}: methods that do not use auto aggregation nor per-pixel rescaling. Furthermore, they do not use manually tuned aggregation weights but simply average the cross-attention maps over all layers.
    \item \textit{Baseline}: methods that do not use auto aggregation nor per-pixel rescaling, but use manually tuned aggregation weights. Since it is hard to tune the weights for other models, the baseline methods only consider SD v1.5.
    \item \textit{Ours}: methods that use both auto aggregation and per-pixel rescaling.
\end{enumerate}
For reference, we still provide the results of some state-of-the-art methods:
\begin{enumerate}[label=(\roman*), leftmargin=25pt]
    \item \textit{Non-DM}: unsupervised semantic segmentation methods that do not involve diffusion models, including MaskCLIP~\cite{DBLP:conf/eccv/ZhouLD22} and ReCO~\cite{DBLP:conf/nips/ShinXA22}.
    \item \textit{Pre-Trained DM}: training-free diffusion segmentors, including DiffSegmentor~\cite{DBLP:journals/corr/abs-2309-02773}, MaskDiffusion~\cite{DBLP:journals/access/KawanoA24a}, and FTTM~\cite{xiao2023text}.
\end{enumerate}
Note that these SOTAs have relatively different settings, especially in how prompts are designed.
This might make direct comparison of these reported results less reliable.

\textbf{Implementation Details.}
It is common practice to rely on external object detectors to modulate prompts for training-free diffusion segmentors, due to the limit of per-token re-normalization. (See \Cref{sec:app-detector}.)
We are using GPT-4o for this purpose and use the same prompts across our own settings.
For alignment test with the prompts used by previous studies, see \Cref{sec:app-alignment}.


\begin{figure*}[t]
  \centering
  \includegraphics[width=.85\linewidth]{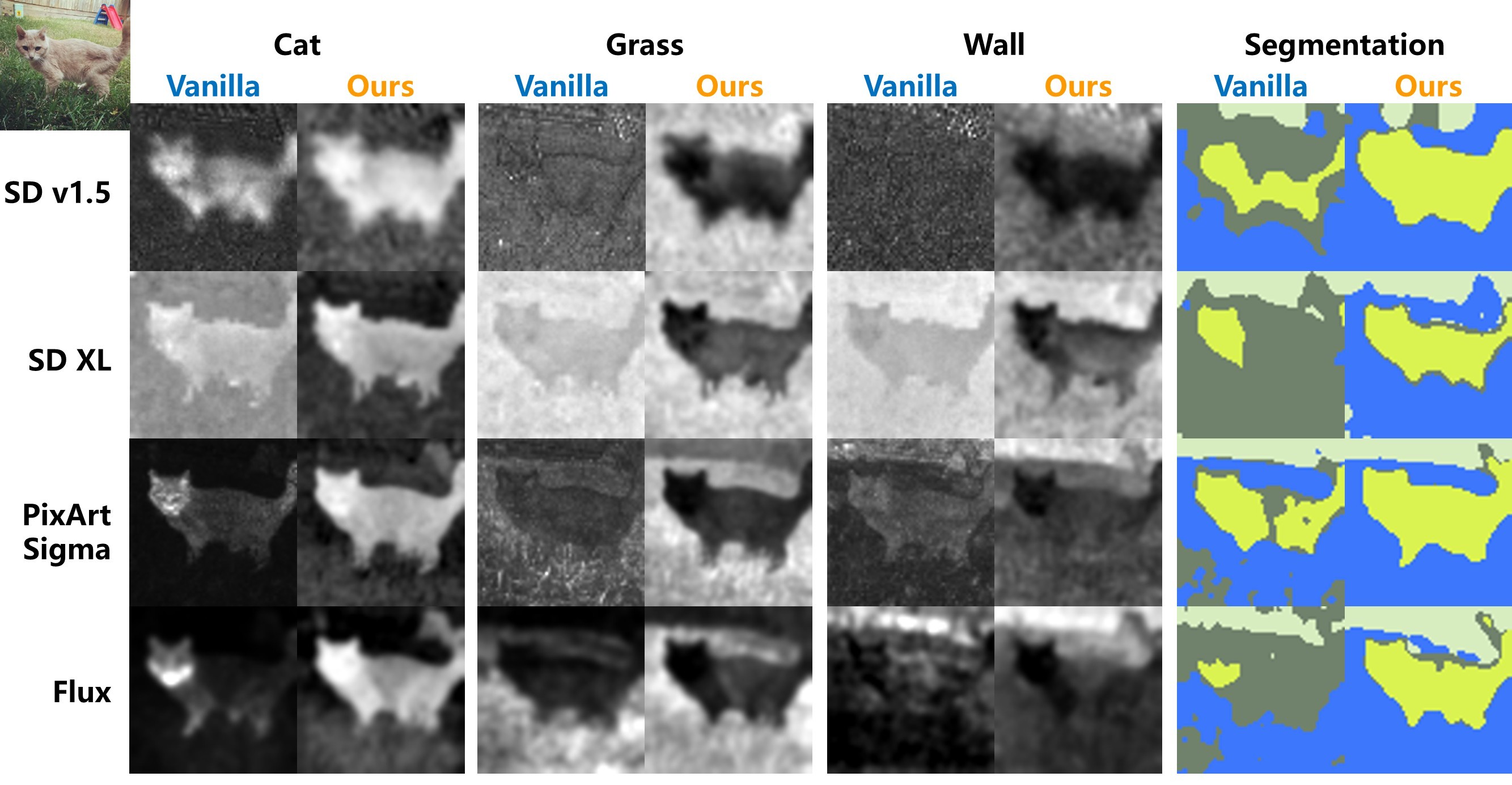}
  \caption{Visualization of per-class attention maps and the final segmentation using \textit{Vanilla} and \textit{Ours}. The input image is as shown in the top-left corner, and the prompt is ``a cat on grass before wall''.
  }
  \label{fig:qualitative}
\end{figure*}

\subsection{Main Results}
From the results in \Cref{tab:exp-benchmark}, we can observe that:
\begin{enumerate}[label=(\roman*), leftmargin=25pt]
    \item \textit{Baseline} methods are generally better than their \textit{Vanilla} counterparts, confirming the benefits of manually tuned aggregation weights. Additionally, \textit{Baseline} with SD v1.5 is sometimes even better than \textit{Vanilla} with SD XL and PixArt-Sigma, despite the two models being much stronger, which is exactly the phenomenon that motivated our study.
    \item \textit{Ours} methods not only achieve much higher performance than their \textit{Baseline} and \textit{Vanilla} counterparts but also manage to make the three stronger models significantly surpass SD v1.5. \textbf{It shows that our method has succeeded in achieving our initial goal of improving the ability to scale with the generative power}.
    \item \textit{Ours} methods outperform \textit{Pre-Trained DM} and \textit{Non-DM} SOTA methods on all datasets, as additional evidence supporting the effectiveness of our method.
\end{enumerate}
For the efficiency test, please refer to \Cref{sec:app-efficiency}.

\subsection{Ablation Study}
\begin{table}[t]
\centering
\caption{Ablation results evaluated using mIoU$\uparrow$ on Pascal VOC 2012 dataset, to evaluate the contribution of auto aggregation and per-pixel rescaling. The best result is marked as \textorange{\textbf{bold}}.
}
\label{tab:exp-ablation}
\begin{tabular}{@{}ccccc@{}}
\toprule
Head & Layer  & Rescaling & SD v1.5 & SD XL            \\ \midrule
{Vanilla} & {Vanilla} & {Vanilla} & 44.3 & 51.1              \\
{Vanilla} & \textbf{Manual} & {Vanilla} & 51.1 & -       \\ \midrule
\textbf{Ours} & {Vanilla} & {Vanilla} & 44.8 & 56.1         \\
{Vanilla} & \textbf{Ours}   & {Vanilla} & 52.1 & 51.3       \\
{Vanilla} & {Vanilla} & \textbf{Ours}      & 52.6 & 51.4    \\ \midrule
\textbf{Ours} & \textbf{Ours}   & \textbf{Ours}      & \textorange{\textbf{60.7}} & \textorange{\textbf{65.6}} \\ \bottomrule
\end{tabular}
\end{table}

\begin{table*}[t]
\centering
\caption{Experimental results as the integration of an advanced generative technique, S-CFG~\cite{S-CFG}, evaluated using FID$\downarrow$ and CLIP$\uparrow$. The top row shows the values of CFG strength, which is a hyperparameter. The best and runner-up are \textorange{\textbf{bold}} and \textblue{\underline{underlined}}.
}
\label{tab:exp-fid}
\begin{tabular}{@{}llcccccccccc@{}}
\toprule
\multicolumn{2}{c}{\multirow{2}{*}{Method}} &
  \multicolumn{2}{c}{CFG=2.0} &
  \multicolumn{2}{c}{CFG=3.0} &
  \multicolumn{2}{c}{CFG=5.0} &
  \multicolumn{2}{c}{CFG=7.5} &
  \multicolumn{2}{c}{CFG=10.0} \\
\multicolumn{2}{c}{} &
  FID &
  CLIP &
  FID &
  CLIP &
  FID &
  CLIP &
  FID &
  CLIP &
  FID &
  CLIP \\ \midrule
\multirow{3}{*}{SD v1.5} &
  CFG &
  37.84 &
  29.84 &
  24.22 &
  30.78 &
  19.27 &
  31.34 &
  18.98 &
  31.53 &
  19.38 &
  31.60 \\
 &
  S-CFG &
  \textblue{\underline{36.45}} &
  \textblue{\underline{29.95}} &
  \textblue{\underline{23.58}} &
  \textblue{\underline{30.84}} &
  \textblue{\underline{19.15}} &
  \textblue{\underline{31.35}} &
  \textblue{\underline{18.90}} &
  \textblue{\underline{31.53}} &
  \textblue{\underline{19.38}} &
  \textblue{\underline{31.60}} \\
 &
  Ours+S-CFG &
  \textorange{\textbf{35.23}} &
  \textorange{\textbf{30.05}} &
  \textorange{\textbf{22.80}} &
  \textorange{\textbf{30.93}} &
  \textorange{\textbf{18.82}} &
  \textorange{\textbf{31.42}} &
  \textorange{\textbf{18.84}} &
  \textorange{\textbf{31.57}} &
  \textorange{\textbf{19.35}} &
  \textorange{\textbf{31.62}} \\
\bottomrule
\end{tabular}
\end{table*}

To evaluate the individual contribution of auto aggregation and per-pixel rescaling, we have conducted an ablation study, with the results listed in \Cref{tab:exp-ablation}.
In this study, we experiment with head-wise and layer-wise parts of auto aggregation, as well as the entire per-pixel rescaling technique.
From the results, we can observe that:
\begin{enumerate}[label=(\roman*), leftmargin=25pt]
    \item Both auto aggregation and per-pixel rescaling, and both head- and layer-wise parts of auto aggregation, contribute to the performance gain. It shows that all components of GoCA are necessary.
    \item Our layer-wise auto aggregation method has a comparable effect to manually tuned layer weights, showing its effectiveness in layer aggregation.
    \item When combined together, the full GoCA can further improve the segmentation quality.
\end{enumerate}
For a more detailed ablation study, where the similarity metric, timestep, and dense feature selection for computing the pseudo attention map are ablated, see \Cref{sec:app-ablation}.

\subsection{Qualitative Study}
We show per-class attention maps and the final segmentation in \Cref{fig:qualitative}. For more visualization, including a failure case, please refer to \Cref{sec:app-vis}.
From the visualization results, we can observe that:
\begin{enumerate}[label=(\roman*), leftmargin=25pt]
    \item Our method provides higher-quality class maps, especially for background objects like ``grass'' and ``wall''.
    \item With our method, stronger diffusion models have stronger segmentation performance. For example, PixArt-Sigma and Flux successfully segment the wooden wall, while the other models are less capable of this.
    \item Since diffusion model interpretation~\cite{daam} also mainly relies on the observation of attention visualization, this qualitative study also serves as a demonstration of GoCA's effectiveness in model interpretation applications.
\end{enumerate}

We also provide a possible explanation of why GoCA is especially effective for background.
It mainly comes from per-pixel rescaling counteracting semantic special token dominance in background regions.
Specifically, we observe that the semantic special token tends to have higher scores in background regions and lower scores in foreground regions, possibly because foreground regions, as salient regions, require more information from other tokens.
As a result, in background regions where background tokens should have larger scores, their scores are adversarially reduced by the semantic special token, and vice versa.
Consequently, the attention maps of background tokens become less discriminative, as shown in ``grass'' and ``wall'' columns of \Cref{fig:qualitative}.
Our per-pixel rescaling, on the other hand, can greatly enhance the quality of background attention maps by excluding the impact of semantic special tokens, leading to the special improvement of background regions in both tasks.

\begin{figure}[t]
  \centering
  \includegraphics[width=.9\linewidth]{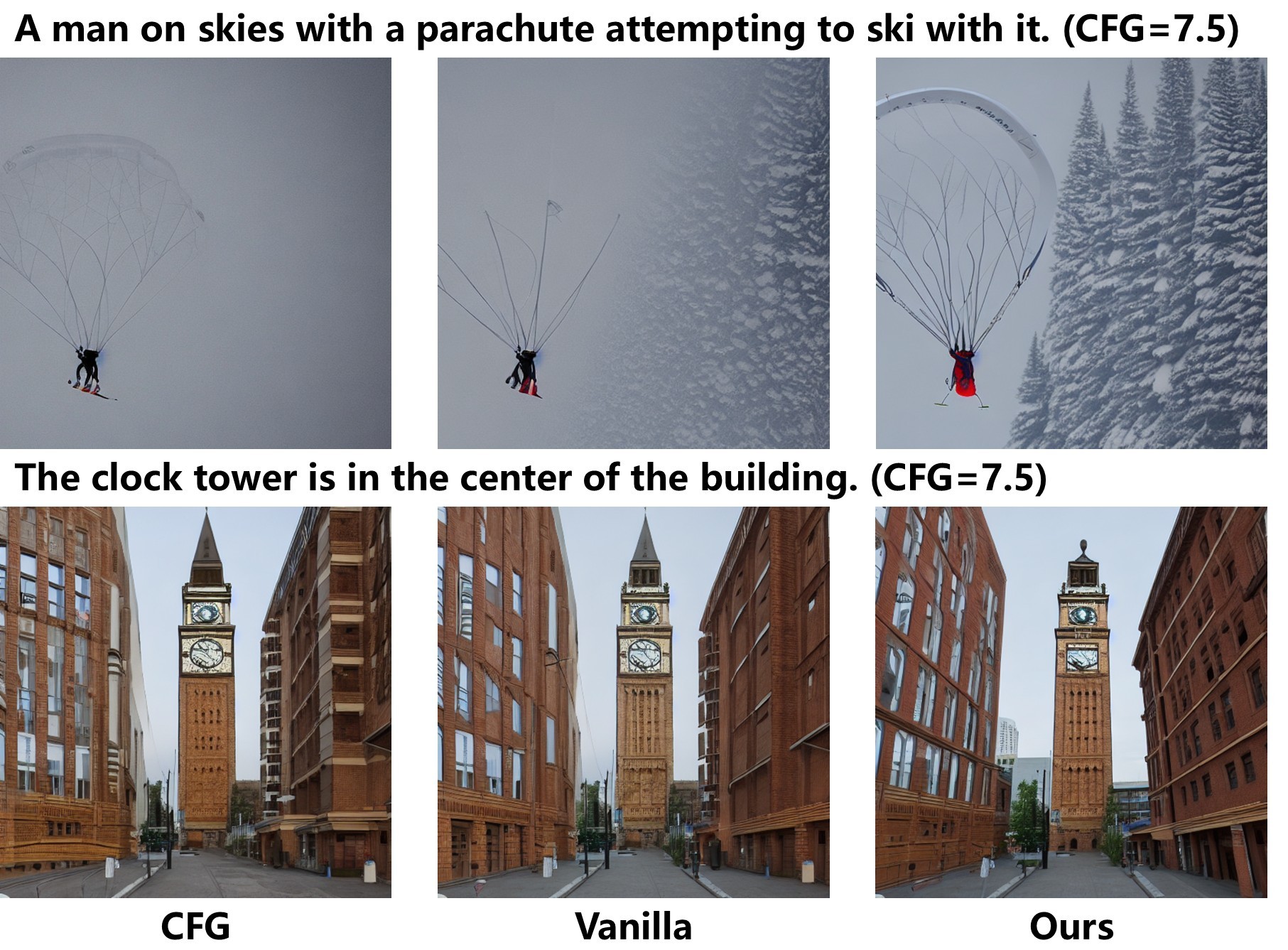}
  \caption{Generated images with SD v1.5. While \textit{Vanilla} improves the background quality over \textit{CFG}, \textit{Ours} further enhances it.
  }
  \label{fig:exp-generation}
\end{figure}

\subsection{Generation Experiment}
We have argued that training-free diffusion segmentors are useful not only in standard segmentation scenarios, but also as an integration in other techniques that rely on training-free segmentation.
To further support this argument, we have chosen a technique called S-CFG~\cite{S-CFG}, targeted at improving the generation quality of diffusion models.
It basically assigns different CFG scales to different pixels, according to the segmentation results yielded by a training-free diffusion segmentor.

The experiment is conducted on COCO-30k dataset.
This dataset actually means randomly selecting 30k image-caption pairs from the validation set of COCO 2014~\cite{coco-object}.
We use the selected set in T2IBenchmark~\cite{boomb0omT2IBenchmark}.
We use FID (Fréchet inception distance) score~\cite{dowson1982frechet} and CLIP score~\cite{DBLP:conf/emnlp/HesselHFBC21} as the metric.
For evaluated methods, we have \textit{Ours+S-CFG}, \textit{S-CFG}, and a setting that does not use S-CFG at all (\textit{CFG}).

The quantitative results are shown in \Cref{tab:exp-fid}.
We also show some samples in \Cref{fig:exp-generation} and \Cref{sec:app-img}, and a significance experiment with three seeds under $CFG=5.0$ in \Cref{sec:app-fid-std}.
From the quantitative results, we find that \textit{Ours+S-CFG} is consistently the best.
It shows that our GoCA is beneficial for advanced generative techniques that utilize training-free diffusion segmentors as their inner components.
The samples also support the observation, where \textit{Ours} yields images with the best background among the three, complying with the qualitative results in \Cref{fig:qualitative}.
This generative integration scenario clearly demonstrates training-free diffusion segmentors' unique value with their inherent modal alignment capability.

\section{Conclusion}
This study is motivated by the observation that current training-free diffusion segmenters face challenges in achieving higher performance with more powerful diffusion models.
To this end, we identify two gaps between cross-attention maps and semantic correlation.
Furthermore, we devise auto aggregation and per-pixel rescaling techniques, correspondingly, helping stronger diffusion models surpass their previous counterparts.

Despite the improvement, our method remains limited to the semantic segmentation task, as other training-free diffusion segmentors do.
Therefore, a promising direction for future work would be to extend the concept of training-free diffusion segmenters to more discriminative tasks beyond segmentation, such as depth estimation or object detection.
This would help training-free diffusion discriminators achieve broader applicability and alleviate the reliance on external object detectors when performing segmentation.

\section*{Acknowledgments}
This work was supported
in part by National Natural Science Foundation of China: 62525212, U23B2051, U2541229, 62411540034, 62236008, 62441232, 62521007, 62502500, and U21B2038,
in part by Youth Innovation Promotion Association CAS,
in part by the Strategic Priority Research Program of the Chinese Academy of Sciences (Grant No. XDB0680201), Ningbo Science and Technology Innovation 2025 Major Project (2025Z027),
in part by the project ZR2025ZD01 supported by Shandong Provincial Natural Science Foundation,
in part by the China National  Postdoctoral Program for Innovative Talents under Grant BX20240384,
in part by Beijing Natural Science Foundation under Grant No. L252144,
in part by General Program of the  Chinese Postdoctoral Science Foundation under Grant No. 2025M771558,
in part by the Beijing Major Science and Technology Project under Contract No. Z251100008125059,
and in part by Beijing Academy of Artificial Intelligence (BAAI).

{
    \small
    \bibliographystyle{ieeenat_fullname}
    \bibliography{main}

@String(CVPR= {IEEE Conf. Comput. Vis. Pattern Recog.})

@String(ECCV= {Eur. Conf. Comput. Vis.})

@String(CVPR  = {CVPR})

@String(ECCV  = {ECCV})

@inproceedings{zhang2023tale,
  author       = {Junyi Zhang and
                  Charles Herrmann and
                  Junhwa Hur and
                  Luisa Polania Cabrera and
                  Varun Jampani and
                  Deqing Sun and
                  Ming{-}Hsuan Yang},
  title        = {A Tale of Two Features: Stable Diffusion Complements {DINO} for Zero-Shot
                  Semantic Correspondence},
  booktitle    = {Annual Conference
                  on Neural Information Processing Systems},
  year         = {2023},
  pages        = {1--15}
}

@inproceedings{xu2023open,
  author       = {Jiarui Xu and
                  Sifei Liu and
                  Arash Vahdat and
                  Wonmin Byeon and
                  Xiaolong Wang and
                  Shalini De Mello},
  title        = {Open-Vocabulary Panoptic Segmentation with Text-to-Image Diffusion
                  Models},
  booktitle    = {{IEEE/CVF} Conference on Computer Vision and Pattern Recognition},
  pages        = {2955--2966},
  year         = {2023},
}

@inproceedings{zhao2023unleashing,
  author       = {Wenliang Zhao and
                  Yongming Rao and
                  Zuyan Liu and
                  Benlin Liu and
                  Jie Zhou and
                  Jiwen Lu},
  title        = {Unleashing Text-to-Image Diffusion Models for Visual Perception},
  booktitle    = {{IEEE/CVF} International Conference on Computer Vision},
  pages        = {5706--5716},
  year         = {2023},
}

@inproceedings{ho2020denoising,
  author       = {Jonathan Ho and
                  Ajay Jain and
                  Pieter Abbeel},
  title        = {Denoising Diffusion Probabilistic Models},
  booktitle    = {Annual Conference
                  on Neural Information Processing Systems},
  year         = {2020},
  pages        = {1--12}
}

@inproceedings{rombach2022high,
  author       = {Robin Rombach and
                  Andreas Blattmann and
                  Dominik Lorenz and
                  Patrick Esser and
                  Bj{\"{o}}rn Ommer},
  title        = {High-Resolution Image Synthesis with Latent Diffusion Models},
  booktitle    = {{IEEE/CVF} Conference on Computer Vision and Pattern Recognition},
  pages        = {10674--10685},
  year         = {2022},
}

@article{podell2023sdxl,
  author       = {Dustin Podell and
                  Zion English and
                  Kyle Lacey and
                  Andreas Blattmann and
                  Tim Dockhorn and
                  Jonas M{\"{u}}ller and
                  Joe Penna and
                  Robin Rombach},
  title        = {{SDXL:} Improving Latent Diffusion Models for High-Resolution Image
                  Synthesis},
  journal      = {CoRR},
  volume       = {abs/2307.01952},
  year         = {2023},
  pages        = {1--21}
}

@inproceedings{yang2023diffusion,
  author       = {Xingyi Yang and
                  Xinchao Wang},
  title        = {Diffusion Model as Representation Learner},
  booktitle    = {{IEEE/CVF} International Conference on Computer Vision},
  pages        = {18892--18903},
  year         = {2023},
}

@article{xiao2023text,
  author       = {Changming Xiao and
                  Qi Yang and
                  Feng Zhou and
                  Changshui Zhang},
  title        = {From text to mask: Localizing entities using the attention of text-to-image
                  diffusion models},
  journal      = {Neurocomputing},
  volume       = {610},
  pages        = {128437},
  year         = {2024},
}

@inproceedings{luo2023dhf,
  author       = {Grace Luo and
                  Lisa Dunlap and
                  Dong Huk Park and
                  Aleksander Holynski and
                  Trevor Darrell},
  title        = {Diffusion Hyperfeatures: Searching Through Time and Space for Semantic
                  Correspondence},
  booktitle    = {Annual Conference
                  on Neural Information Processing Systems},
  year         = {2023},
  pages        = {1--11}
}

@article{hao2020brief,
  author       = {Shijie Hao and
                  Yuan Zhou and
                  Yanrong Guo},
  title        = {A Brief Survey on Semantic Segmentation with Deep Learning},
  journal      = {Neurocomputing},
  volume       = {406},
  pages        = {302--321},
  year         = {2020},
}

@inproceedings{baranchuk2021label,
  author       = {Dmitry Baranchuk and
                  Andrey Voynov and
                  Ivan Rubachev and
                  Valentin Khrulkov and
                  Artem Babenko},
  title        = {Label-Efficient Semantic Segmentation with Diffusion Models},
  booktitle    = {The Tenth International Conference on Learning Representations},
  year         = {2022},
  pages        = {1--15}
}

@inproceedings{dhariwal2021guideddiffusion,
  author       = {Prafulla Dhariwal and
                  Alexander Quinn Nichol},
  title        = {Diffusion Models Beat GANs on Image Synthesis},
  booktitle    = {Annual Conference
                  on Neural Information Processing Systems},
  pages        = {8780--8794},
  year         = {2021},
}

@inproceedings{vaswani2017attention,
  author       = {Ashish Vaswani and
                  Noam Shazeer and
                  Niki Parmar and
                  Jakob Uszkoreit and
                  Llion Jones and
                  Aidan N. Gomez and
                  Lukasz Kaiser and
                  Illia Polosukhin},
  title        = {Attention is All you Need},
  booktitle    = {Annual Conference
                  on Neural Information Processing Systems},
  pages        = {5998--6008},
  year         = {2017},
}

@inproceedings{Cordts16city,
  author       = {Marius Cordts and
                  Mohamed Omran and
                  Sebastian Ramos and
                  Timo Rehfeld and
                  Markus Enzweiler and
                  Rodrigo Benenson and
                  Uwe Franke and
                  Stefan Roth and
                  Bernt Schiele},
  title        = {The Cityscapes Dataset for Semantic Urban Scene Understanding},
  booktitle    = {{IEEE} Conference on Computer Vision and Pattern Recognition},
  pages        = {3213--3223},
  year         = {2016},
}

@inproceedings{Zhou17ade20k,
  author       = {Bolei Zhou and
                  Hang Zhao and
                  Xavier Puig and
                  Sanja Fidler and
                  Adela Barriuso and
                  Antonio Torralba},
  title        = {Scene Parsing through {ADE20K} Dataset},
  booktitle    = {{IEEE} Conference on Computer Vision and Pattern Recognition},
  pages        = {5122--5130},
  year         = {2017},
}

@inproceedings{DBLP:conf/cvpr/DuttMM24,
  author       = {Niladri Shekhar Dutt and
                  Sanjeev Muralikrishnan and
                  Niloy J. Mitra},
  title        = {Diffusion 3D Features (Diff3F) Decorating Untextured Shapes with Distilled
                  Semantic Features},
  booktitle    = {{IEEE/CVF} Conference on Computer Vision and Pattern Recognition},
  pages        = {4494--4504},
  year         = {2024},
}

@inproceedings{frick2024diffseg,
  title={DiffSeg: Towards Detecting Diffusion-Based Inpainting Attacks Using Multi-Feature Segmentation},
  author={Frick, Raphael Antonius and Steinebach, Martin},
  booktitle={{IEEE/CVF} Conference on Computer Vision and Pattern Recognition},
  pages={3802--3808},
  year={2024}
}

@inproceedings{pixart-sigma,
  author       = {Junsong Chen and
                  Chongjian Ge and
                  Enze Xie and
                  Yue Wu and
                  Lewei Yao and
                  Xiaozhe Ren and
                  Zhongdao Wang and
                  Ping Luo and
                  Huchuan Lu and
                  Zhenguo Li},
  title        = {PIXART-{\(\Sigma\)}: Weak-to-Strong Training of Diffusion Transformer
                  for 4K Text-to-Image Generation},
  booktitle    = {Computer Vision 18th European Conference},
  volume       = {15090},
  pages        = {74--91},
  year         = {2024},
}

@misc{pascal-voc-2012,
	author = "Everingham, M. and Van~Gool, L. and Williams, C. K. I. and Winn, J. and Zisserman, A.",
	title = "The {PASCAL} {V}isual {O}bject {C}lasses {C}hallenge 2012 {(VOC2012)} {R}esults",
	howpublished = "http://www.pascal-network.org/challenges/VOC/voc2012/workshop/index.html",
        year={2012}}

@inproceedings{DBLP:journals/corr/abs-2406-02842,
 author = {Couairon, Paul and Shukor, Mustafa and HAUGEARD, Jean-Emmanuel and Cord, Matthieu and THOME, Nicolas},
 booktitle = {Advances in Neural Information Processing Systems},
 pages = {13548--13578},
 title = {DiffCut: Catalyzing Zero-Shot Semantic Segmentation with Diffusion Features and Recursive Normalized Cut},
 volume = {37},
 year = {2024}
}

@article{DBLP:journals/corr/abs-2309-02773,
  author       = {Jinglong Wang and
                  Xiawei Li and
                  Jing Zhang and
                  Qingyuan Xu and
                  Qin Zhou and
                  Qian Yu and
                  Lu Sheng and
                  Dong Xu},
  title        = {Diffusion Model is Secretly a Training-free Open Vocabulary Semantic
                  Segmenter},
  journal      = {CoRR},
  volume       = {abs/2309.02773},
  year         = {2023},
  pages        = {1--9}
}

@article{DBLP:journals/access/KawanoA24a,
  author       = {Yasufumi Kawano and
                  Yoshimitsu Aoki},
  title        = {MaskDiffusion: Exploiting Pre-Trained Diffusion Models for Semantic
                  Segmentation},
  journal      = {{IEEE} Access},
  volume       = {12},
  pages        = {127283--127293},
  year         = {2024},
}

@inproceedings{daam,
  author       = {Raphael Tang and
                  Linqing Liu and
                  Akshat Pandey and
                  Zhiying Jiang and
                  Gefei Yang and
                  Karun Kumar and
                  Pontus Stenetorp and
                  Jimmy Lin and
                  Ferhan Ture},
  title        = {What the {DAAM:} Interpreting Stable Diffusion Using Cross Attention},
  booktitle    = {Proceedings of the 61st Annual Meeting of the Association for Computational
                  Linguistics},
  pages        = {5644--5659},
  year         = {2023},
}

@inproceedings{S-CFG,
  author       = {Dazhong Shen and
                  Guanglu Song and
                  Zeyue Xue and
                  Fu{-}Yun Wang and
                  Yu Liu},
  title        = {Rethinking the Spatial Inconsistency in Classifier-Free Diffusion
                  Guidance},
  booktitle    = {{IEEE/CVF} Conference on Computer Vision and Pattern Recognition},
  pages        = {9370--9379},
  year         = {2024},
}

@inproceedings{my-SDXL,
  author       = {Benyuan Meng and
                  Qianqian Xu and
                  Zitai Wang and
                  Xiaochun Cao and
                  Qingming Huang},
  title        = {Not All Diffusion Model Activations Have Been Evaluated as Discriminative
                  Features},
  booktitle    = {Annual Conference on Neural Information Processing Systems},
  year         = {2024},
  pages        = {55141--55177},
}

@inproceedings{my-GATE,
  author       = {Benyuan Meng and
                  Qianqian Xu and
                  Zitai Wang and
                  Zhiyong Yang and
                  Xiaochun Cao and
                  Qingming Huang},
  title        = {Suppress Content Shift: Better Diffusion Features via Off-the-Shelf
                  Generation Techniques},
  booktitle    = {Annual Conference on Neural Information Processing Systems},
  year         = {2024},
  pages        = {18910--18939},
}

@InProceedings{pascal-context,
 author       = {Roozbeh Mottaghi and Xianjie Chen and Xiaobai Liu and Nam-Gyu Cho and Seong-Whan Lee and Sanja Fidler and Raquel Urtasun and Alan Yuille},
 title        = {The Role of Context for Object Detection and Semantic Segmentation in the Wild},
 booktitle    = {IEEE Conference on Computer Vision and Pattern Recognition (CVPR)},
 year         = {2014},
}

@inproceedings{coco-object,
  author       = {Tsung{-}Yi Lin and
                  Michael Maire and
                  Serge J. Belongie and
                  James Hays and
                  Pietro Perona and
                  Deva Ramanan and
                  Piotr Doll{\'{a}}r and
                  C. Lawrence Zitnick},
  title        = {Microsoft {COCO:} Common Objects in Context},
  booktitle    = {Computer Vision - {ECCV} 2014 - 13th European Conference},
  volume       = {8693},
  pages        = {740--755},
  year         = {2014},
}

@misc{boomb0omT2IBenchmark,
  author={Pavlov, I. and Ivanov, A. and Stafievskiy, S.},
  title={{Text-to-Image Benchmark: A benchmark for generative models}},
  howpublished={\url{https://github.com/boomb0om/text2image-benchmark}},
  month={September},
  year={2023},
  note={Version 0.1.0},
}

@inproceedings{DBLP:conf/eccv/ZhouLD22,
  author       = {Chong Zhou and
                  Chen Change Loy and
                  Bo Dai},
  title        = {Extract Free Dense Labels from {CLIP}},
  booktitle    = {Computer Vision - {ECCV} 2022 - 17th European Conference},
  volume       = {13688},
  pages        = {696--712},
  year         = {2022},
}

@inproceedings{DBLP:conf/nips/ShinXA22,
  author       = {Gyungin Shin and
                  Weidi Xie and
                  Samuel Albanie},
  title        = {ReCo: Retrieve and Co-segment for Zero-shot Transfer},
  booktitle    = {Advances in Neural Information Processing Systems 35},
  year         = {2022},
}

@inproceedings{DBLP:conf/iclr/XiaoL0WH24,
  author       = {Jiayu Xiao and
                  Henglei Lv and
                  Liang Li and
                  Shuhui Wang and
                  Qingming Huang},
  title        = {R{\&}B: Region and Boundary Aware Zero-shot Grounded Text-to-image
                  Generation},
  booktitle    = {The Twelfth International Conference on Learning Representations},
  year         = {2024},
}

@inproceedings{DBLP:conf/cvpr/Liu0CJH24,
  author       = {Bingyan Liu and
                  Chengyu Wang and
                  Tingfeng Cao and
                  Kui Jia and
                  Jun Huang},
  title        = {Towards Understanding Cross and Self-Attention in Stable Diffusion
                  for Text-Guided Image Editing},
  booktitle    = {{IEEE/CVF} Conference on Computer Vision and Pattern Recognition},
  pages        = {7817--7826},
  year         = {2024},
}

@article{dowson1982frechet,
  title={The Fr{\'e}chet distance between multivariate normal distributions},
  author={Dowson, DC and Landau, BV666017},
  journal={Journal of multivariate analysis},
  volume={12},
  number={3},
  pages={450--455},
  year={1982},
  publisher={Elsevier}
}

@inproceedings{DBLP:conf/emnlp/HesselHFBC21,
  author       = {Jack Hessel and
                  Ari Holtzman and
                  Maxwell Forbes and
                  Ronan Le Bras and
                  Yejin Choi},
  title        = {CLIPScore: {A} Reference-free Evaluation Metric for Image Captioning},
  booktitle    = {Proceedings of the 2021 Conference on Empirical Methods in Natural
                  Language Processing},
  pages        = {7514--7528},
  publisher    = {Association for Computational Linguistics},
  year         = {2021},
}

@article{DBLP:journals/corr/abs-2410-11842,
  author       = {Peng Jin and
                  Bo Zhu and
                  Li Yuan and
                  Shuicheng Yan},
  title        = {MoH: Multi-Head Attention as Mixture-of-Head Attention},
  journal      = {CoRR},
  volume       = {abs/2410.11842},
  year         = {2024},
}

@inproceedings{DBLP:conf/aaai/ParkJB25,
  author       = {Joon Hyun Park and
                  Kumju Jo and
                  Sungyong Baik},
  title        = {SeeDiff: Off-the-Shelf Seeded Mask Generation from Diffusion Models},
  booktitle    = {AAAI-25, Sponsored by the Association for the Advancement of Artificial
                  Intelligence},
  pages        = {6406--6415},
  year         = {2025},
}

@article{DBLP:journals/corr/abs-2409-03209,
  author       = {Lin Sun and
                  Jiale Cao and
                  Jin Xie and
                  Fahad Shahbaz Khan and
                  Yanwei Pang},
  title        = {iSeg: An Iterative Refinement-based Framework for Training-free Segmentation},
  journal      = {CoRR},
  volume       = {abs/2409.03209},
  year         = {2024},
}

@inproceedings{DBLP:conf/cvpr/ZhangYM24,
  author       = {Xiao Zhang and
                  David Yunis and
                  Michael Maire},
  title        = {Deciphering 'What' and 'Where' Visual Pathways from Spectral Clustering
                  of Layer-Distributed Neural Representations},
  booktitle    = {{IEEE/CVF} Conference on Computer Vision and Pattern Recognition},
  pages        = {4165--4175},
  year         = {2024},
}

@misc{flux2024,
    author={Black Forest Labs},
    title={FLUX},
    year={2024},
    howpublished={\url{https://github.com/black-forest-labs/flux}},
}

@article{DBLP:journals/corr/abs-2502-04320,
  author       = {Alec Helbling and
                  Tuna Han Salih Meral and
                  Benjamin Hoover and
                  Pinar Yanardag and
                  Duen Horng Chau},
  title        = {ConceptAttention: Diffusion Transformers Learn Highly Interpretable
                  Features},
  journal      = {CoRR},
  volume       = {abs/2502.04320},
  year         = {2025},
}

@inproceedings{DBLP:conf/iclr/XiaoTCHL24,
  author       = {Guangxuan Xiao and
                  Yuandong Tian and
                  Beidi Chen and
                  Song Han and
                  Mike Lewis},
  title        = {Efficient Streaming Language Models with Attention Sinks},
  booktitle    = {International Conference on Learning Representations},
  year         = {2024},
}

@article{DBLP:journals/corr/abs-2505-06708,
  author       = {Zihan Qiu and
                  Zekun Wang and
                  Bo Zheng and
                  Zeyu Huang and
                  Kaiyue Wen and
                  Songlin Yang and
                  Rui Men and
                  Le Yu and
                  Fei Huang and
                  Suozhi Huang and
                  Dayiheng Liu and
                  Jingren Zhou and
                  Junyang Lin},
  title        = {Gated Attention for Large Language Models: Non-linearity, Sparsity,
                  and Attention-Sink-Free},
  journal      = {CoRR},
  volume       = {abs/2505.06708},
  year         = {2025},
}

@inproceedings{han2024aucseg,
  title={Aucseg: Auc-oriented pixel-level long-tail semantic segmentation},
  author={Han, Boyu and Xu, Qianqian and Yang, Zhiyong and Bao, Shilong and Wen, Peisong and Jiang, Yangbangyan and Huang, Qingming},
  booktitle={Advances in Neural Information Processing Systems},
  pages={126863--126907},
  year={2024}
}

@inproceedings{han2025lightfair,
  title={LightFair: Towards an Efficient Alternative for Fair T2I Diffusion via Debiasing Pre-trained Text Encoders},
  author={Han, Boyu and Xu, Qianqian and Bao, Shilong and Yang, Zhiyong and Zi, Kangli and Huang, Qingming},
  booktitle={Advances in Neural Information Processing Systems},
  year={2025}
}

@article{DBLP:journals/pami/WangXYXZCH26,
  author       = {Zitai Wang and
                  Qianqian Xu and
                  Zhiyong Yang and
                  Zhikang Xu and
                  Linchao Zhang and
                  Xiaochun Cao and
                  Qingming Huang},
  title        = {A Unified Perspective for Loss-Oriented Imbalanced Learning via Localization},
  journal      = {{IEEE} Trans. Pattern Anal. Mach. Intell.},
  volume       = {48},
  number       = {1},
  pages        = {639--656},
  year         = {2026},


}

@inproceedings{DBLP:conf/nips/WangX00CH23,
  author       = {Zitai Wang and
                  Qianqian Xu and
                  Zhiyong Yang and
                  Yuan He and
                  Xiaochun Cao and
                  Qingming Huang},
  title        = {A Unified Generalization Analysis of Re-Weighting and Logit-Adjustment
                  for Imbalanced Learning},
  booktitle    = {Advances in Neural Information Processing Systems},
  year         = {2023},
}

@inproceedings{yang2024harnessing,
    title={Harnessing Hierarchical Label Distribution Variations in Test Agnostic Long-tail Recognition}, 
    author={Zhiyong Yang and Qianqian Xu and Zitai Wang and Sicong Li and Boyu Han and Shilong Bao and Xiaochun Cao and Qingming Huang},
    booktitle={International Conference on Machine Learning},
    year={2024}
}

@inproceedings{DBLP:conf/icml/Li00WZCH25,
  author       = {Sicong Li and
                  Qianqian Xu and
                  Zhiyong Yang and
                  Zitai Wang and
                  Linchao Zhang and
                  Xiaochun Cao and
                  Qingming Huang},
  title        = {Focal-SAM: Focal Sharpness-Aware Minimization for Long-Tailed Classification},
  booktitle    = {International Conference on Machine Learning},
  year         = {2025},
}
}

\newpage
\vspace*{\fill}
\newpage
\appendix
\section{Reproduction Details}
\label{sec:app-reproduction}

In \Cref{tab:exp-benchmark}, we show the reproduced results of FTTM~\cite{xiao2023text} instead of its originally reported results.
This is because, in the original paper, a different setting is adopted.
To be specific, Xiao \etal~\cite{xiao2023text} used the predictions given by a training-free diffusion segmentor as image-segmentation pairs to train another supervised neural network to yield the final result.
In our setting, however, the predictions given by training-free diffusion segmentors are directly adopted as the final result.
As a consequence, we choose to reproduce the FTTM method in our setting.

\section{Visualization of Auto Weights}
\begin{figure}[t]
  \centering
  \includegraphics[width=.95\linewidth]{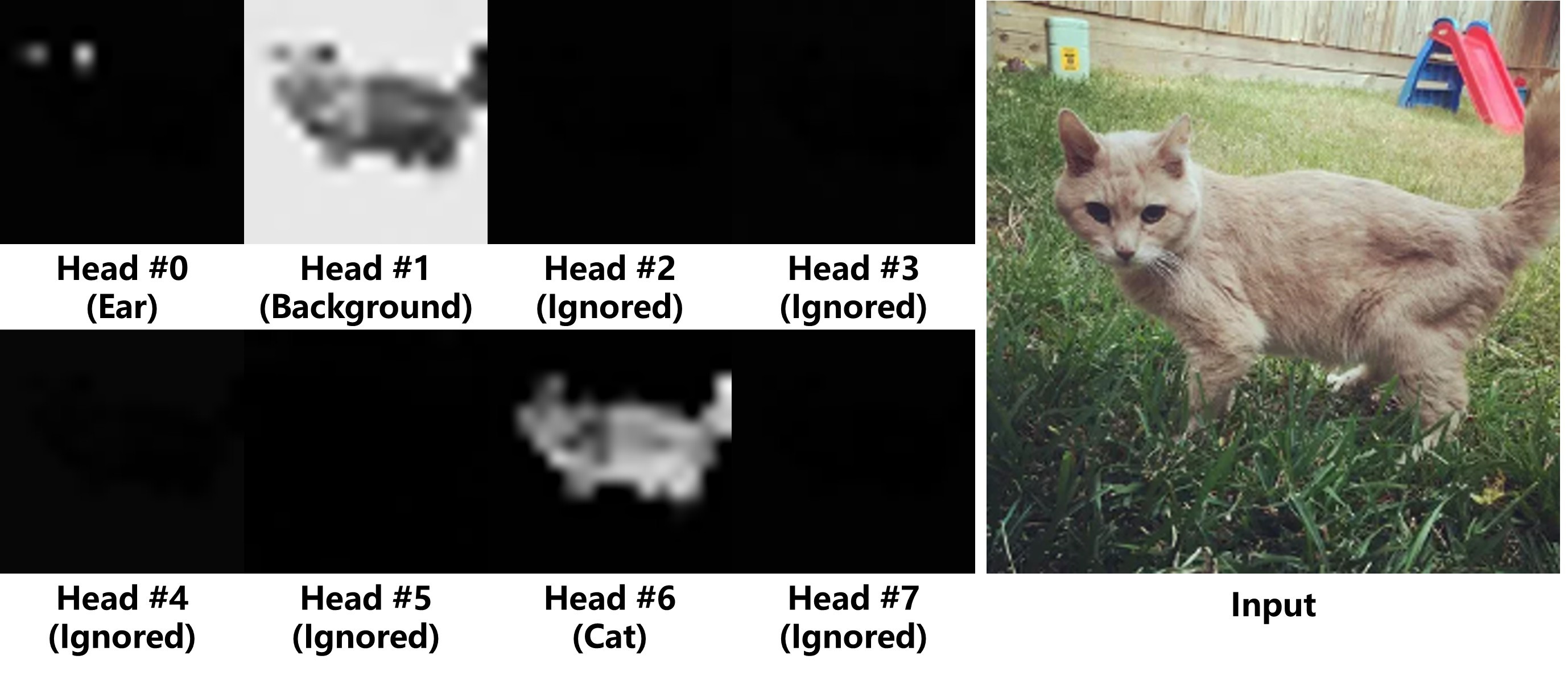}
  \caption{Head-wise auto-aggregation weights in a certain layer of SD v1.5.
  }
  \label{fig:app-weight-head}
\end{figure}

\begin{figure}[t]
  \centering
  \includegraphics[width=.95\linewidth]{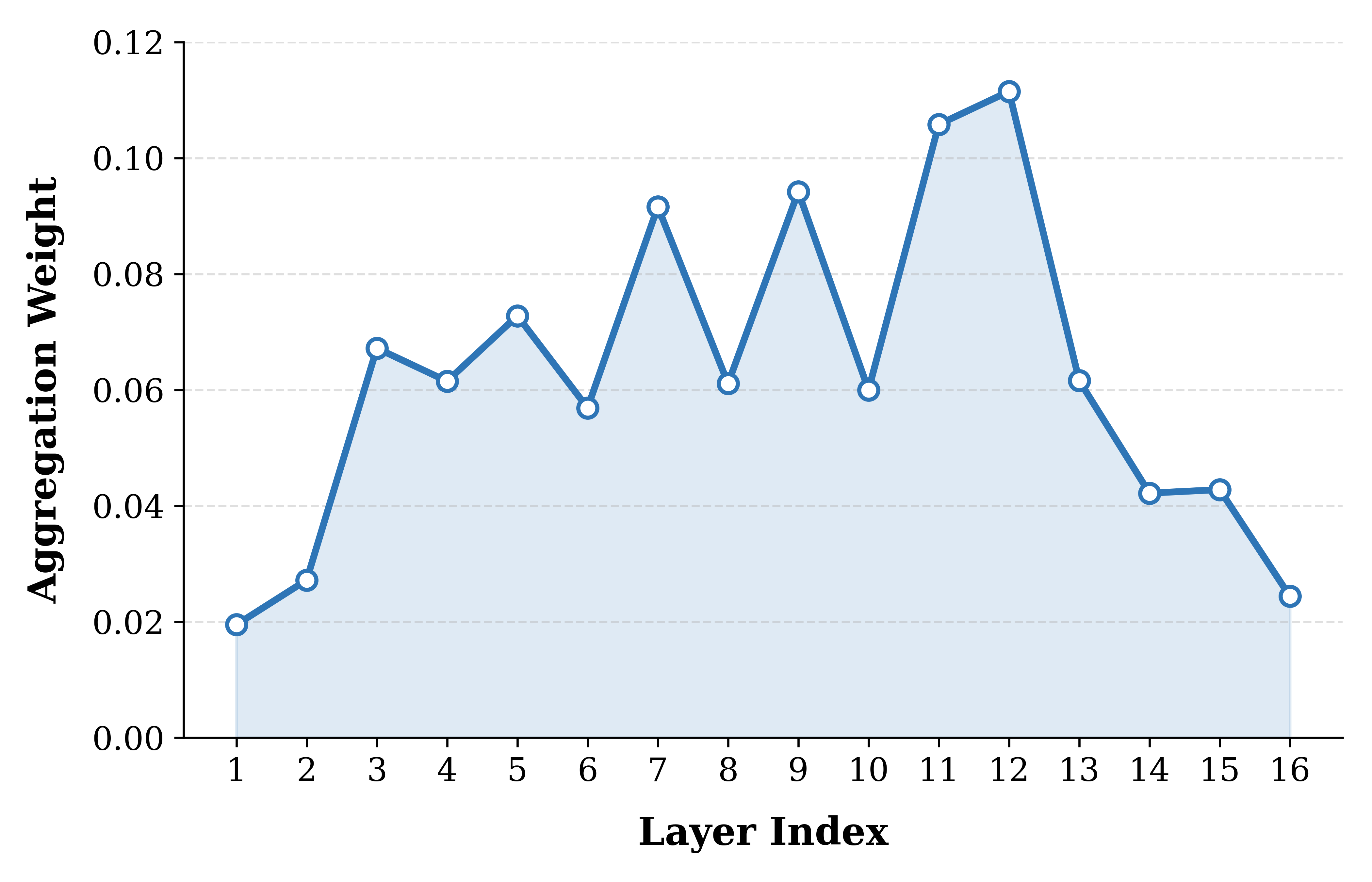}
  \caption{Layer-wise auto-aggregation weights of SD v1.5, using the cat input as shown in \Cref{fig:app-weight-head}.
  }
  \label{fig:app-weight-layer}
\end{figure}

We show example head-wise and layer-wise weights derived with auto-aggregation in \Cref{fig:app-weight-head} and \Cref{fig:app-weight-layer}.
Both examples are based on the SD v1.5 model.
The head-wise weights derived with auto-aggregation are per-pixel weights, meaning that each pixel has its own head weights, allowing fine-grained aggregation.
The layer-wise weights, in contrast, are shared by all pixels.

\section{Complexity Analysis}
\label{sec:app-efficiency}
The computation overhead introduced by all components of GoCA is nearly negligible.
Specifically, head-wise aggregation is linear to the number of heads and very minor, layer-wise aggregation is linear to the number of layers, and per-pixel re-scaling uses constant time.

To further strengthen our claim of negligible overhead, we also actually track the running time in \Cref{tab:app-time}.
From the results, we can observe that the overhead introduced by GoCA is indeed small.

\begin{table}[]
\centering
\caption{Running time comparison on VOC 2012 dataset with the SD v1.5 model.}
\label{tab:app-time}
\begin{tabular}{@{}lc@{}}
\toprule
Method      & Running Time (s) \\ \midrule
SD v1.5 Vanilla  & 753.53           \\
SD v1.5 Baseline & 750.09           \\
SD v1.5          & 767.73           \\ \bottomrule
\end{tabular}
\end{table}

\section{Reliance on External Object Detectors}
\label{sec:app-detector}
Due to the use of per-token re-normalization, training-free diffusion segmentors need to rely on external object detectors.
To be specific, even when an object does not exist in the given image and has very low scores, its corresponding scores will still be normalized to $[0,1]$ with per-token re-normalization.
As a result, it is likely that some pixels will be assigned to this class nonetheless.
Therefore, it is usually considered necessary to first filter non-existent objects from the prompts to obtain favorable segmentation results~\cite{DBLP:journals/corr/abs-2309-02773, DBLP:journals/access/KawanoA24a, xiao2023text}.
We follow this practice in this study, but see this as a drawback to overcome in future work.
Moreover, since we are using a different external object detector from previous studies, we conduct an additional prompt alignment experiment in \Cref{sec:app-alignment} to show fairness.

\section{Additional Ablation}
\label{sec:app-ablation}
\subsection{Similarity Metrics}
We use dot-product similarity for both head- and layer-wise aggregation.
In this section, we ablate this choice by comparing with other similarity metrics, with results in \Cref{tab:app-ablation-sim}.
For head-wise aggregation, we have tried $l_2$-norm and cosine similarity beyond dot-product similarity.
For layer-wise aggregation, we have tried the MSE loss and continuous IoU between self-attention maps beyond dot-product similarity.
We can observe that dot-product is not always the best choice across different models.
However, it is relatively robust, yielding satisfying results in different cases.
Therefore, we choose dot-product similarity in the method design.

\begin{table*}[]
\centering
\caption{Ablation results to compare different similarity metrics for head- and layer-wise aggregation.}
\label{tab:app-ablation-sim}
\begin{tabular}{@{}lllc@{}}
\toprule
Model                         & Head Metric & Layer Metric & mIoU (Context) \\ \midrule
\multirow{5}{*}{SD v1.5}      & dot-product & dot-product  & \textblue{\underline{40.4}}           \\
                              & l2-norm     & dot-product  & \textorange{\textbf{42.0}}           \\
                              & cosine      & dot-product  & \textorange{\textbf{42.0}}           \\
                              & dot-product & mse          & 33.9           \\
                              & dot-product & iou-like     & 39.2           \\ \midrule
\multirow{5}{*}{SD XL}        & dot-product & dot-product  & \textorange{\textbf{42.3}}           \\
                              & l2-norm     & dot-product  & 40.3           \\
                              & cosine      & dot-product  & 40.3           \\
                              & dot-product & mse          & \textblue{\underline{42.1}}           \\
                              & dot-product & iou-like     & 41.9           \\ \midrule
\multirow{5}{*}{PixArt-Sigma} & dot-product & dot-product  & 43.2           \\
                              & l2-norm     & dot-product  & 43.2           \\
                              & cosine      & dot-product  & 41.2           \\
                              & dot-product & mse          & \textorange{\textbf{44.1}}           \\
                              & dot-product & iou-like     & \textblue{\underline{43.4}}           \\ \bottomrule
\end{tabular}
\end{table*}

\subsection{Timesteps}
In \Cref{tab:app-ablation-t}, we compare the performance on different timesteps.
We can observe that GoCA is not very sensitive to timesteps, and it is relatively easy to find the optimal timestep using grid search.

\begin{table*}[]
\centering
\caption{Ablation results to compare different timesteps for feature extraction.}
\label{tab:app-ablation-t}
\begin{tabular}{@{}llc@{}}
\toprule
Model                    & Timestep & \multicolumn{1}{l}{IoU (Context)} \\ \midrule
\multirow{4}{*}{SD v1.5} & 50       & 40.1                              \\
                         & 100      & \textorange{\textbf{40.4}}                              \\
                         & 150      & \textblue{\underline{40.3}}                              \\
                         & 200      & 39.7                              \\ \bottomrule
\end{tabular}
\end{table*}

\subsection{Dense Feature}
In layer-wise auto aggregation, we propose to use a dense feature to calculate the pseudo self-attention.
In \Cref{tab:app-ablation-layer}, we compare the performance with different layers used to extract the dense features.
In the table, we use the layer notation and reported feature quality as in \cite{my-SDXL}.
We can observe that GoCA is not very sensitive to dense feature selection, and it is a generic principle to find a dense feature with relatively high quality.

\begin{table*}[]
\centering
\caption{Ablation results to compare different dense feature layers for layer-wise auto aggregation.}
\label{tab:app-ablation-layer}
\begin{tabular}{@{}llccc@{}}
\toprule
Model                    & Dense Layer                          & Dense Feature Quality (mIoU) & Resolution & IoU (Context) \\ \midrule
\multirow{4}{*}{SD v1.5} & up-level3-repeat1-vit-block0-self-q  & 31.64                        & 64       & 39.4          \\
                         & up-level2-repeat1-vit-block0-cross-q & 53.58                        & 32       & \textorange{\textbf{40.4}}          \\
                         & up-level2-repeat2-vit-block0-cross-q & 46.24                        & 32       & \textorange{\textbf{40.4}}          \\
                         & up-level1-repeat1-vit-block0-cross-q & 49.04                        & 16       & \textblue{\underline{40.3}}          \\ \bottomrule
\end{tabular}
\end{table*}

\subsection{Prompt Alignment}
\label{sec:app-alignment}
In our experiment, we use GPT-4o as the external object detector.
This is because we find GoCA especially effective at segmenting background objects, but usually, background objects are not listed as the classes of common datasets, making it hard to harness this improvement.
Additionally, our per-pixel rescaling step requires the comparison between multiple objects, but there are cases where only one class is annotated for an image, making GoCA not applicable.
Hence, we utilize the open-vocabulary capability of an LLM to detect both in-vocabulary and out-of-vocabulary objects in the images.
Afterward, instead of the vanilla background threshold approach usually adopted by previous methods, we combine it with the segmentation results of these background objects.
To be specific, we set the background score of a pixel as the maximum value of the threshold and all background object scores at this pixel.
This makes GoCA applicable to single-object samples and gives GoCA results a tighter contour.

Note that GPT-4o does not yield prompts so perfect that the improvement of GoCA totally comes from GPT-4o.
For example, GPT-4o does not always follow the format instruction, potentially introducing errors into the prompt parsing process.
Additionally, GPT-4o and the annotation standard of a specific dataset may name the same object differently or perceive the same scene differently, leading to degradation.
In fact, we find that, though GPT-4o prompts can lead to improvement with GoCA, SOTA methods tend to degenerate with them.
This is demonstrated by the results in \Cref{tab:app-ablation-alignment} (excluding the final row), where we change the prompts of DiffSegmenter~\cite{DBLP:journals/corr/abs-2309-02773} to what we use for GoCA.
In this table, \textit{GPT-4o w/o background} indicates a setting where we discard the background objects detected by GPT-4o and only use the in-vocabulary ones.
\textit{GPT-4o w/ background} is where we adopt a similar strategy as with GoCA to set the background scores dynamically.
For both settings, there is significant degradation of segmentation capability.
This demonstrates that although GoCA indeed relies on GPT-4o to gain some improvement and become more widely applicable, the enhancement does not totally come from GPT-4o.

\begin{table}[]
\centering
\caption{Ablation results where we modify the implementation of DiffSegmenter~\cite{DBLP:journals/corr/abs-2309-02773}, a SOTA method, and experiment on Context dataset.}
\label{tab:app-ablation-alignment}
\begin{tabular}{@{}lc@{}}
\toprule
Method                & Context \\ \midrule
Original              & \textorange{\textbf{60.4}}    \\
GPT-4o w/o background & \textblue{\underline{47.1}}    \\
GPT-4o w/ background  & 41.3    \\
SD XL                 & 43.2    \\ \bottomrule
\end{tabular}
\end{table}

\subsection{More Powerful Model with SOTA}
In the main experiment, we do not apply \textit{Baseline} to stronger diffusion models, but only SD v1.5.
This is because manually tuning the layer-wise weights is not practical for more complex models.
Nevertheless, experimenting with more powerful models under stronger settings may still be desirable.
Hence, we propose a workaround: we switch the model in DiffSegmenter to SD XL in the bottom row of \Cref{tab:app-ablation-alignment}.
We use a weight of 0.7 for level 0 attention maps and a weight of 0.3 for level 1 attention maps (level 1 has a higher resolution).
We can see that there is a significant performance drop, showing that SOTA methods can not effectively work with more powerful diffusion models.

\section{Significance Experiment for Generation}
\label{sec:app-fid-std}
In \Cref{tab:exp-fid}, the gain of FID scores is consistent but not very large.
To better support the claim that GoCA is beneficial for S-CFG as its integrated component, we select one setting ($CFG=5.0$) to conduct the same experiment under two additional random seeds.
With three sets of results in total, we report the mean and standard deviation values in \Cref{tab:app-fid}.
As observed, the gain of \textit{S-CFG} over \textit{CFG} and \textit{Ours+S-CFG} over \textit{S-CFG} is consistent across all seeds, showing the benefit of GoCA as an integration in other techniques where training-free diffusion segmentors are needed.

\begin{table*}[]
\centering
\caption{Experimental results with three different seeds under $CFG=5.0$ setting for FID and CLIP score evaluation.}
\label{tab:app-fid}
\begin{tabular}{@{}lcccccccc@{}}
\toprule
Method & FID (1) & FID (2) & FID (3) & FID (AVG)  & CLIP (1) & CLIP (2) & CLIP (3) & CLIP (AVG) \\ \midrule
CFG    & 19.27   & 19.97   & 21.09   & 20.11±0.92 & 31.34    & 31.12    & 30.95    & 31.13±0.20 \\
S-CFG  & \textblue{\underline{19.15}}   & \textblue{\underline{18.52}}   & \textblue{\underline{20.41}}   & \textblue{\underline{19.36±0.96}} & \textblue{\underline{31.35}}    & \textblue{\underline{31.26}}    & \textblue{\underline{31.16}}    & \textblue{\underline{31.25±0.10}} \\
Ours+S-CFG   & \textorange{\textbf{18.82}}   & \textorange{\textbf{18.01}}   & \textorange{\textbf{20.32}}   & \textorange{\textbf{19.05±1.17}} & \textorange{\textbf{31.42}}    & \textorange{\textbf{31.28}}    & \textorange{\textbf{31.16}}    & \textorange{\textbf{31.28±0.13}} \\ \bottomrule
\end{tabular}
\end{table*}

\section{Exact Feature Extraction Specification}
\textbf{Stable Diffusion v1.5:}
Extracting features at $t=100$.
Using cross-attention query activation from 2nd ViT, 3rd upsampling resolution to calculate the pseudo global attention.
Treating ``\textless sos\textgreater'' (start-of-sentence) as semantic special tokens.

\textbf{Stable Diffusion v1.5} (\textit{Baseline}):
For this setting, we extract cross-attention maps from layer 5 to 12, which are then averaged uniformly.

\textbf{Stable Diffusion XL:}
Extracting features at $t=100$.
Using cross-attention query activation from 1st basic block, 1st ViT, 2nd upsampling resolution to calculate the pseudo global attention.
This model has two text encoders, but their outputs are concatenated along the channels dimension, so we can treat them as a single text encoder when extracting attention maps.
This model is a bit special about semantic special tokens: it has a dedicated pooled embedding to serve as the global embedding, and all ``\textless sos\textgreater'' tokens have exactly zero scores.
However, in this model, stop word tokens have much higher scores, somehow serving the role of semantic special tokens.
Hence, our decision to exclude both semantic special tokens and stop word tokens can work with Stable Diffusion XL.

\textbf{PixArt-Sigma:}
Extracting features at $t=100$.
Using cross-attention query activation from 14th basic block to calculate the pseudo global attention.
Treating ``\textless eos\textgreater'' (end-of-sentence) as semantic special tokens.

\textbf{Flux:}
Extracting features at $t=150$.
Using attention output activation from 28th basic block to calculate the pseudo global attention.
Specially, in the MMDiT~\cite{flux2024} architecture, there are both image-query-text-key/value attention sections and text-query-image-key/value attention sections.
We only extract attention maps from where the image serves as the query and text serves as the key/value, to be aligned with U-Net and DiT settings.
Additionally, this model has two text encoders, but only one of them produces the per-token embedding, so we can treat this model as having only one text encoder when extracting attention maps.
Treating ``\textless eos\textgreater'' (end-of-sentence) as semantic special tokens.

Besides the model-specific implementation, we simply treat all tokens that are neither semantic special tokens nor objects of interest (content word tokens) as stop word tokens.
In this way, we have a generic way to define the token categories.

There are also other diffusion models where the outputs of multiple text encoders are concatenated along the sequence length dimension.
We can extract corresponding tokens from all text encoders.
For example, if ``cat'' is encoded by both text encoders, there will be two tokens for this word, and we simply average their scores.
This is the same as how we deal with regular multi-word objects, such as ``potted plants''.

\section{Additional Generation Results}
\label{sec:app-img}

\begin{figure*}[t]
  \centering
  \includegraphics[width=.75\linewidth]{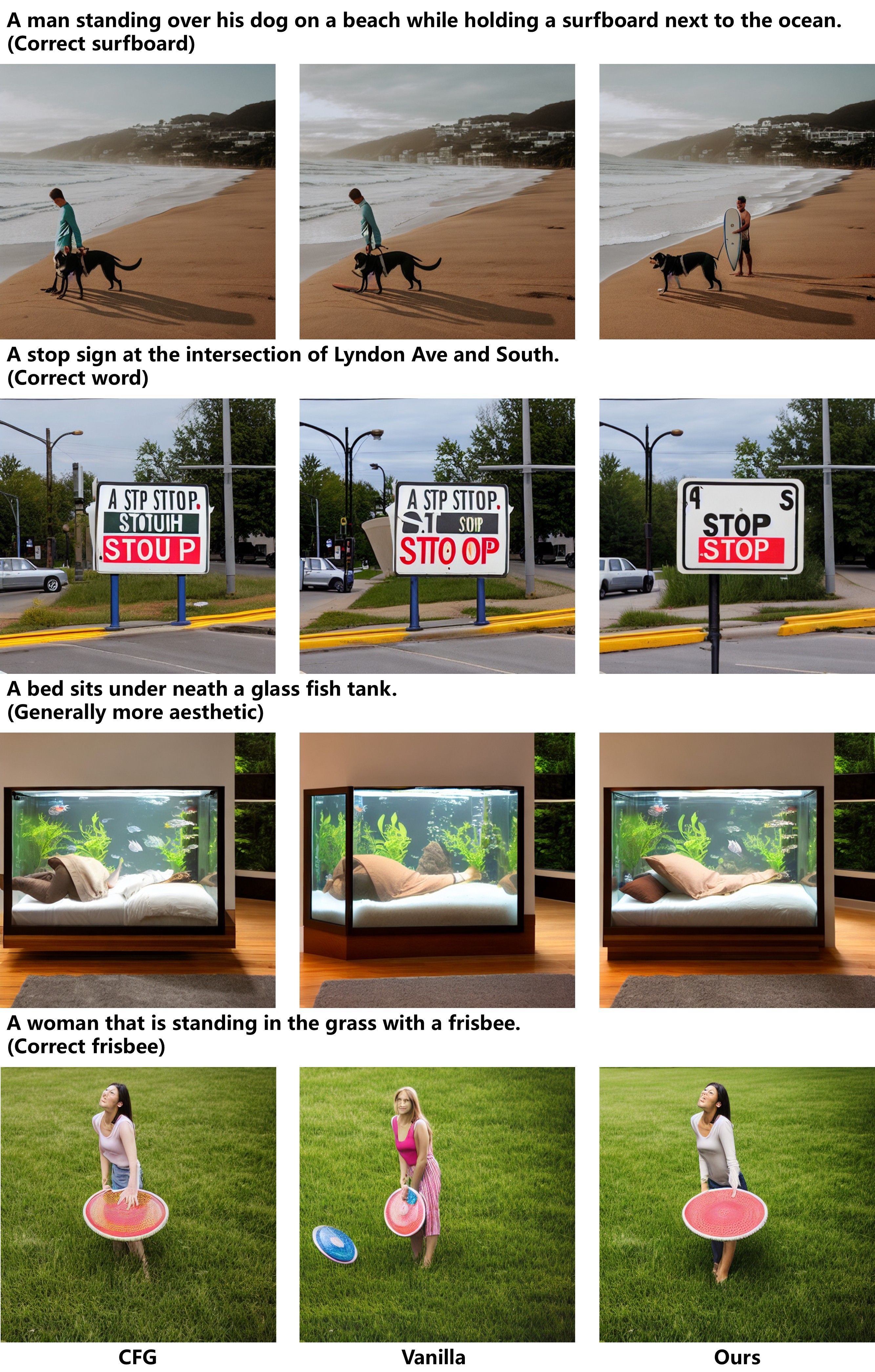}
  \caption{Generated images with SD v1.5 using $CFG=7.5$ setting. Corresponding prompts and why \textit{Ours} is considered better are annotated above each row of images.
  }
  \label{fig:app-generation-15}
\end{figure*}

We show additional SD v1.5 generation results in \Cref{fig:app-generation-15}.
\textit{Ours} can enhance the generation quality more than \textit{Vanilla}, showing the effectiveness of GoCA.

\section{Additional Segmentation Visualization}
\label{sec:app-vis}

\begin{figure*}[t]
  \centering
  \includegraphics[width=.95\linewidth]{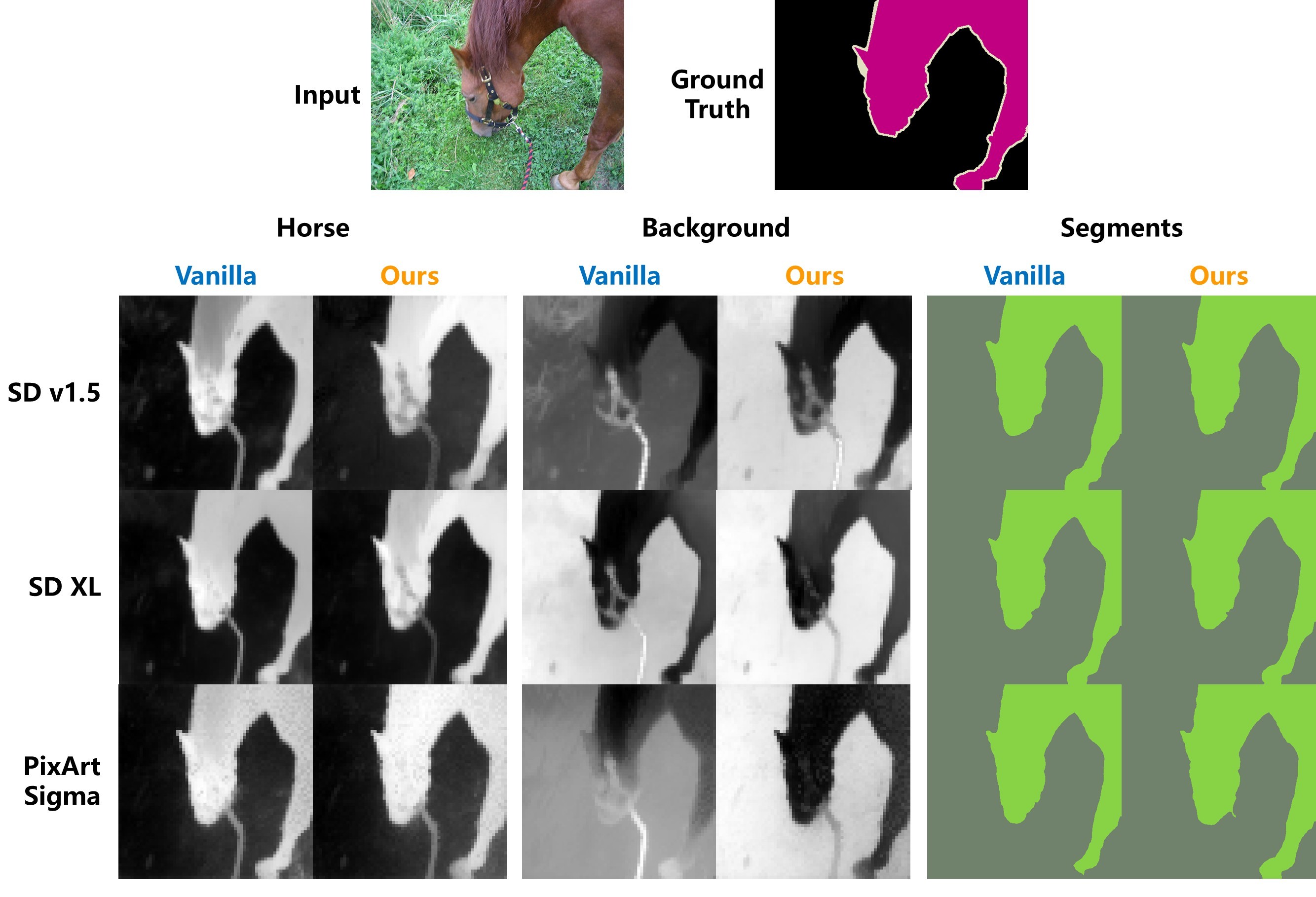}
  \caption{In a simple scene, though there are quality gaps in per-class maps, the final segmentation maps are almost identical.
  }
  \label{fig:app-seg1}
\end{figure*}

\begin{figure*}[t]
  \centering
  \includegraphics[width=.95\linewidth]{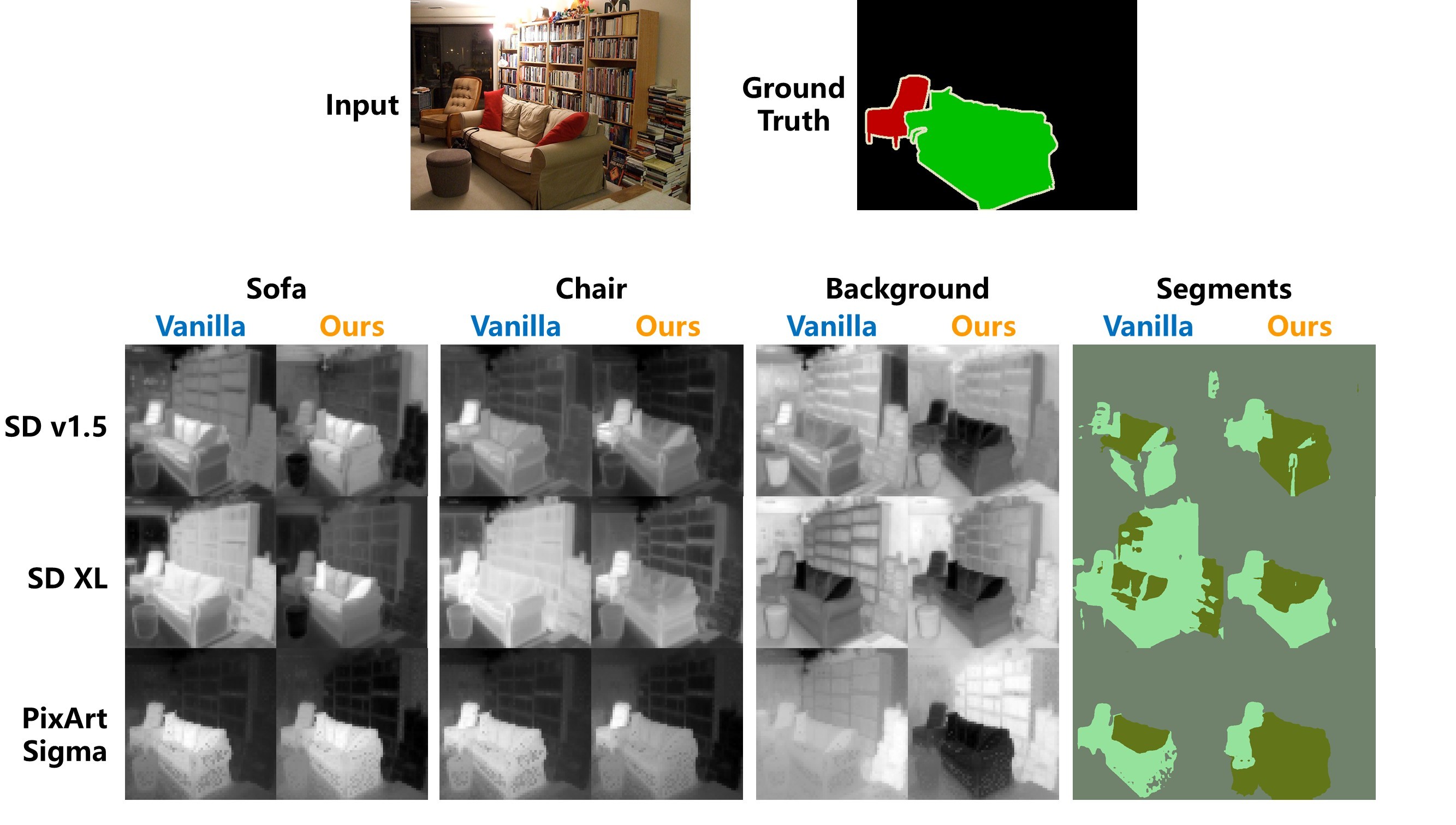}
  \caption{In a moderately complex scene, \textit{Ours} gains significant quality improvement, especially in the background regions.
  }
  \label{fig:app-seg2}
\end{figure*}

\begin{figure*}[t]
  \centering
  \includegraphics[width=.95\linewidth]{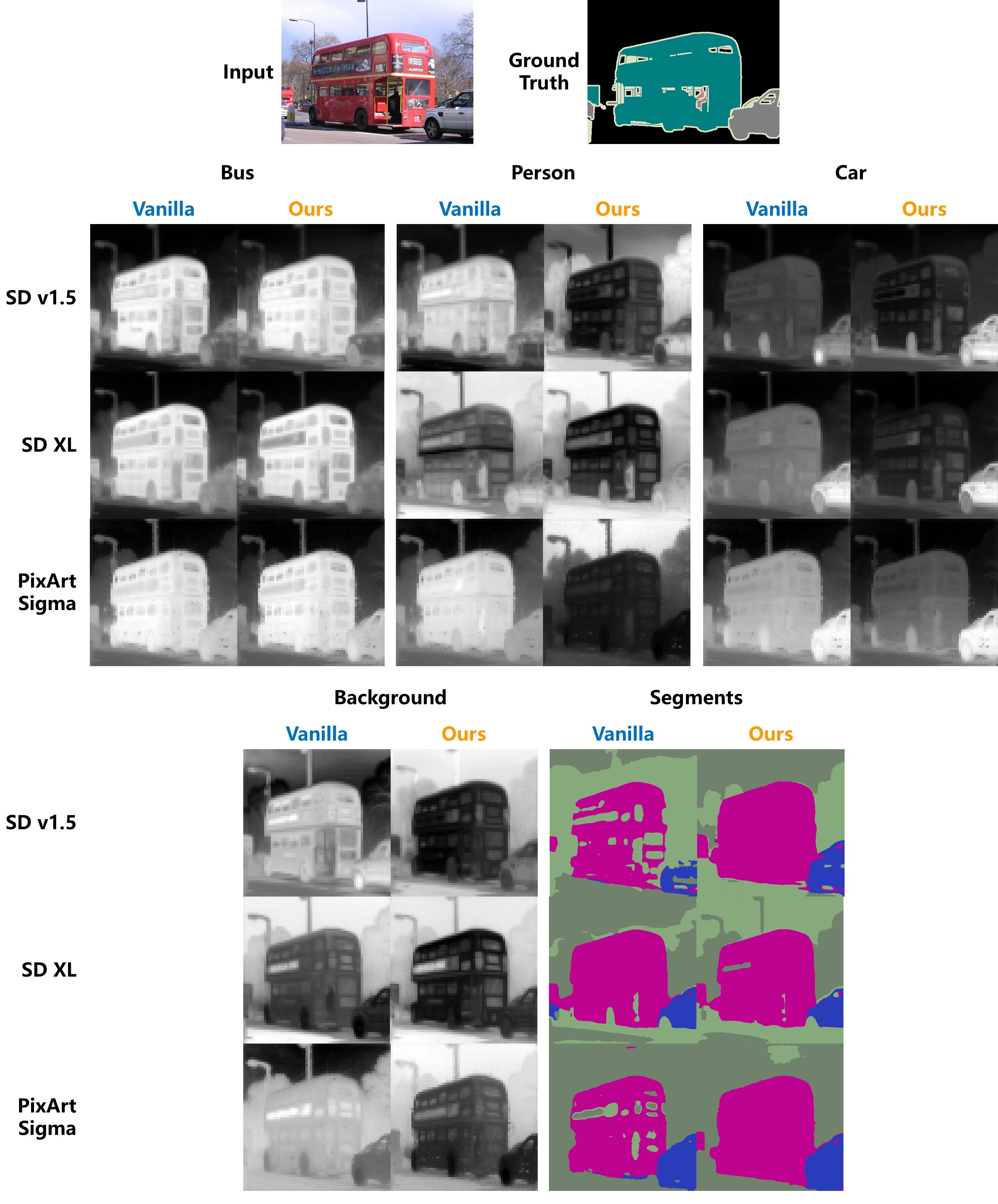}
  \caption{All models and both methods fail to capture the small person object in the complex scene, as latent diffusion models compress the original image with VAE. Nevertheless, we can still observe obvious quality enhancement in the segmentation of background regions, covering the sky, trees, and the road.
  }
  \label{fig:app-seg3}
\end{figure*}

We provide three groups of segmentation visualization in three increasingly more complex scenes, in \Cref{fig:app-seg1}, \Cref{fig:app-seg2}, and \Cref{fig:app-seg3}.
Notably, we can observe a failure pattern from \Cref{fig:app-seg3}:
when an object is very small with respect to how many pixels it takes, the segmentation for this object tends to fail.
This might be because the VAE compresses the original image, or because small objects are long-tail knowledge during training~\cite{DBLP:journals/pami/WangXYXZCH26, DBLP:conf/nips/WangX00CH23, yang2024harnessing, DBLP:conf/icml/Li00WZCH25}.

\end{document}